\def\year{2020}\relax
\documentclass[letterpaper]{article} % DO NOT CHANGE THIS
\usepackage{aaai20}  % DO NOT CHANGE THIS
\usepackage{times}  % DO NOT CHANGE THIS
\usepackage{helvet} % DO NOT CHANGE THIS
\usepackage{courier}  % DO NOT CHANGE THIS
\usepackage[hyphens]{url}  % DO NOT CHANGE THIS
\usepackage{graphicx} % DO NOT CHANGE THIS
\usepackage{amsmath}
\usepackage{amssymb}
\usepackage[ruled,vlined]{algorithm2e}

\usepackage{graphicx}  % DO NOT CHANGE THIS
\newcommand{\citet}[1]{\citeauthor{#1} \shortcite{#1}} \newcommand{\citep}{\cite} 
\DeclareMathOperator*{\argmax}{arg\,max}

\title{Few-Shot Bayesian Imitation Learning with Logical Program Policies}
\author{
Tom Silver, Kelsey R. Allen, Alex K. Lew, Leslie Kaelbling, Josh Tenenbaum\\
Massachusetts Institute of Technology \\
\{tslvr, krallen, alexlew, lpk, jbt\}@mit.edu
}
\begin{document}

\maketitle

\begin{abstract}
Humans can learn many novel tasks from a very small number (1--5) of demonstrations, in stark contrast to the data requirements of nearly tabula rasa deep learning methods. We propose an expressive class of policies, a strong but general prior, and a learning algorithm that, together, can learn interesting policies from very few examples.   We represent policies as logical combinations of programs drawn from a domain-specific language (DSL), define a prior over policies with a probabilistic grammar, and derive an approximate Bayesian inference algorithm to learn policies from demonstrations. In experiments, we study six strategy games played on a 2D grid with one shared DSL. After a few demonstrations of each game, the inferred policies generalize to new game instances that differ substantially from the demonstrations. Our policy learning is 20--1,000x more data efficient than convolutional and fully convolutional policy learning and many orders of magnitude more computationally efficient than vanilla program induction. We argue that the proposed method is an apt choice for tasks that have scarce training data and feature significant, structured variation between task instances.
\end{abstract}

\section{Introduction}

\begin{figure}[h]
    \centering
    \includegraphics[width=0.45\textwidth]{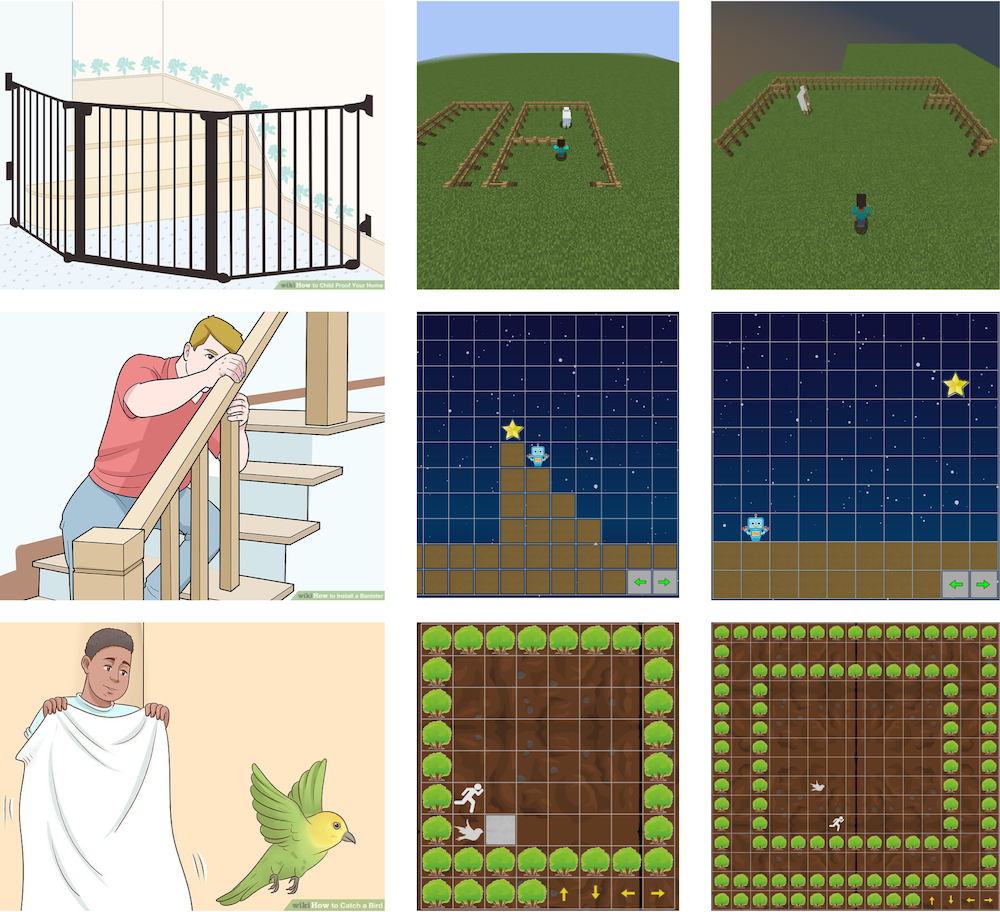}
    \caption{People can learn strategies for an enormous variety of tasks from one or a few demonstrations, e.g., ``gate off an area,'' ``build stairs,'' or ``catch a bird'' (left). We propose a policy class and learning algorithm for similarly data-efficient imitation learning. Given 1--5 demos of tasks like ``Fence In,'' ``Reach for the Star,'' and ``Chase'' (middle), we learn policies that generalize substantially (right).}
    \label{fig:illustrative_tasks}
\end{figure}

People are remarkably good at learning and generalizing strategies for everyday tasks, like ironing a shirt or brewing a cup of coffee, from one or a few demonstrations. Websites like \url{WikiHow.com} and \url{LifeHacker.com} are filled with thousands of ``how-to'' guides for tasks that are hard to solve by pure reasoning or trial and error alone, but easy to learn and generalize from just one illustrated demo (Figure \ref{fig:illustrative_tasks}). We are interested in designing artificial agents with the same few-shot imitation learning capabilities.

A common approach to imitation learning is behavior cloning (BC), in which demonstrations are used as supervision to directly train a policy. BC is often thought to be too prone to overfitting to generalize from very little data. Indeed, we find that neural network policies trained with BC are suspectible to severe overfitting in our experiments. However, we argue that this failure is due not to BC in general, but rather, to an underconstrained policy class and a weak prior.

More structured policies with strong Occam's razor priors can be found in two lines of work: logical and relational (policy) learning \cite{dvzeroski1998relational,natarajan2011imitation}, and program (policy) synthesis \cite{wingate2013compositional,sun2018neural}. Policies expressed in predicate logic are easy to learn, but difficult to scale, since each possible predicate must be hand-engineered by the researcher. Programmatic policies can be automatically generated by searching a small domain-specific language (DSL), but learning even moderately sophisticated policies can require an untenably large search in program space.

We propose Logical Program Policies (\texttt{LPP}): an expressive, structured, and efficiently learnable policy class that combines the strengths of logical and programmatic policies. Our first main idea is to consider policies that have logical ``top level'' structure and programmatic feature detectors (predicates) at the ``bottom level.'' The feature detectors are expressions in a domain-specific language (DSL). By logically combining feature detectors, we can derive an infinitely large, rich policy class from a small DSL. This ``infinite use of finite means'' is in contrast to prior work in relational RL where each feature is individually engineered, making it labor-intensive to apply in complex settings.
%limiting scalability.

Our second main idea is to exploit the logical structure of \texttt{LPP} to obtain an efficient imitation learning algorithm, overcoming the intractability of general program synthesis. To find policies in \texttt{LPP}, we incrementally enumerate feature detectors, apply them to the demonstrations, invoke an off-the-shelf Boolean learning method, and score each candidate policy with a likelihood and prior. What would be an intractable search over full policies is effectively reduced to a manageable search over feature detectors. We thus have an efficient approximate Bayesian inference method for $\mathsf{p}(\pi | \mathcal{D})$, the posterior distribution of policies $\pi$ given demonstrations $\mathcal{D}$.

While \texttt{LPP} and the proposed learning method are agnostic to the particular choice of the DSL and application domain, we focus here on six strategy games which are played on 2D grids of arbitrary size (see Figure \ref{fig:all_games}). In these games, a state consists of an assignment of discrete values to each grid cell and an action is a single grid cell (a ``click'' on the grid). The games are diverse in their transition rules and in the tactics required to win, but the common state and action spaces allow us to build our policies for all six games from one shared, small DSL. 
%Our DSL is inspired by ways that humans focus and shift their attention between cells when analyzing a scene. 
In experiments, we find that policies learned from five or fewer demonstrations can generalize perfectly in all six games. 
In contrast, policies learned as convolutional neural networks fail to generalize, even when domain-specific locality structure is built into the architecture (as in fully convolutional networks \citep{long2015fully}). 
Overall, our experiments suggest that \texttt{LPP} offers an efficient, flexible framework for learning rich, generalizable policies from very little data.

\section{Problem Statement}

In imitation learning, we are given a dataset $\mathcal{D}$ of expert trajectories $(s_0, a_0, ..., s_{T-1}, a_{T-1}, s_T)$ where $s_t \in \mathcal{S}$ are states and $a_t \in \mathcal{A}$ are actions. 
% In this work, we take the state space $\mathcal{S} = \bigcup_{m,n \in \mathbb{N}} V^{mn}$ to contain all grids with $V$-valued cells, and the action space $\mathcal{A} = \mathbb{N} \times \mathbb{N}$ to be (row, column) indices, indicating which cell to click.
%In this work, we assume that states and actions are discrete, but allow for an unbounded number of both. 
We suppose that the trajectories are sampled from a Markov process $\mathcal{M} = (\mathcal{S}, \mathcal{A}, T, \mathcal{G})$, with transition distribution $T(s' \mid s, a)$ and goal states $\mathcal{G} \subset \mathcal{S}$, and that actions are sampled from an expert policy  $\pi^* : \mathcal{S} \times \mathcal{A} \to [0, 1]$, where $\pi^*(a \mid s)$ is a state-conditional distribution over actions. For imitation learning, we must specify (1) a hypothesis class of policies $\Pi$ and (2) an algorithm for learning a policy $\pi \in \Pi$ from $\mathcal{D}$ that matches the expert $\pi^*$. We assume that the expert $\pi^*$ is optimal with respect to $\mathcal{M}$, so we report the fraction of trials in which a learned policy $\pi$ reaches goal states in $\mathcal{G}$ from held-out initial states in $\mathcal{M}$ to evaluate performance.

\section{The \texttt{LPP} Policy Class}

We seek a policy class $\Pi$ with a concise parameterization that can be reasonably specified by a human programmer, and for which there is a tractable learning algorithm for recovering $\pi^* \in \Pi$ from demonstrations.

% We seek a policy class $\Pi$ that satisfies the following desiderata: (1) For all expert policies of interest $\pi^*$, there exists a $\pi \in \Pi$ equivalent to $\pi^*$; (2) There is a concise parameterization of $\Pi$ that can be reasonably specified by a human programmer; and (3) There is a tractable learning algorithm for recovering $\pi \in \Pi$ close to $\pi^*$ from demonstrations.

We consider policies that are parameterized by state-action classifiers $h : \mathcal{S} \times \mathcal{A} \rightarrow \{0, 1\}$. When $h(s, a) = 0$, action $a$ will never be taken in state $s$; when $h(s, a) = 1$, $a$ may be taken. This parameterization allows us to handle arbitrarily large action spaces (variable grid sizes). Given $h(s, a)$, we can derive a corresponding policy $\pi(a \mid s)$ that samples $a$ uniformly at random among those $a$ such that $h(s, a) = 1$. In other words, $\pi(a \mid s) \propto h(s, a)$. This stochastic policy formulation reflects the fact that the demonstrator may randomly select among several optimal actions. For completeness, we define $\pi(a \mid s) \propto 1$ if $\forall a, h(s, a) = 0$. Specifying a policy class $\Pi$ thus reduces to specifying a class of functions $\mathcal{H}$ from which to learn an $h$. 

One option for $\mathcal{H}$ is to consider \textit{logical} rules that compute Boolean expressions combining binary features derived from $(s, a)$. Although this enables fast inference using well-understood Boolean learning algorithms, it requires the AI programmer to hand-engineer informative binary features, which will necessarily vary from task to task. Another option is to consider \textit{programmatic} rules: rules that are expressions in some general-purpose DSL for predicates on state-action pairs. In this case, the AI programmer need only specify a small core of primitives for the DSL, from which task-specific policies can be derived during inference. The challenge here is that finding a good policy in the infinitely large class of programs in the DSL is difficult; simple methods like enumeration are much too slow to be useful.
% Logical policies can be efficiently learned, but their scalability is limited, since all of the logical predicates must be individually specified.
% A program-based policy class can be concisely represented with a grammar, but scalability is again limited because learning becomes intractable with even moderately sophisticated languages. 
% Our policy class combines two ideas that have appeared before in the literature. ``Logical'' policies are parameterized by logical rules that classify (state, action) pairs as admissible or not; confronted with a state `s`, the agent chooses an action `a` at random among those actions for which `(s, a)` satisfies the rule. The rule might be expressed, e.g., as a decision tree or logical formula. A benefit of this approach is that there exist several efficient learning algorithms for deriving these decision trees or logical formulae. The disadvantage is that the AI researcher must hand-engineer the logical predicates that occur at each node of the decision tree. On the other hand, ``programmatic'' policies are parameterized by 

We combine the complementary strengths of logical and program-based policies to define the Logical Program Policies (\texttt{LPP}) class. Policies in \texttt{LPP} have a logical ``top level'' and a programmatic ''bottom level.'' The bottom level is comprised of feature detector programs $f : \mathcal{S} \times \mathcal{A} \to \{ 0, 1 \}$. These programs are expressions in a DSL and can include, for example, loops and conditional statements. A feature detector program takes a state $s$ and an action $a$ as input and returns a binary output, which provides one bit of information about whether $a$ should be taken in $s$. % Mention that The same feature-detector program can be run over grids of arbitrary size?
The top level is comprised of a logical formula $h$ over the outputs of the bottom level. Without loss of generality, we can express the formula in disjunctive normal form:
\begin{equation}
\label{eq:classifier}
\begin{split}
h(s, a) \triangleq (f_{1,1}(s, a) \land ... \land f_{1,n_1}(s, a)) \lor ... \\
  \lor (f_{m,1}(s, a) \land ... \land f_{m,n_m}(s, a))
\end{split}
\end{equation}
where the $f$'s are possibly negated. \texttt{LPP} thus includes all policies that correspond to logical formulae over finite subsets of feature detector programs expressed in the DSL.

\section{Imitation Learning as Bayesian Inference}
\label{sec:learning}

We now address the imitation learning problem of finding a policy $\pi$ that fits the expert demonstrations $\mathcal{D}$. Rather than finding a single LPP policy, we will infer a full posterior distribution over policies $\mathsf{p}(\pi \mid \mathcal{D})$. From a Bayesian perspective, maintaining the full posterior is principled; from a practical perspective, the full posterior leads to modest performance gains over a single MAP policy. Once we have inferred $\mathsf{p}(\pi \mid \mathcal{D})$, we will ultimately take MAP actions according to $\argmax_{a \in \mathcal{A}} \mathbb{E}_{\mathsf{p}(\pi \mid \mathcal{D})}[\pi(a \mid s)]$.

% Once we have an approximation $\mathsf{q}$ to the posterior, we can use it to derive a final policy for use at test time:
% \begin{equation*}
% % \begin{split}
%     \pi_*(s) = \argmax_{a \in \mathcal{A}} \mathbb{E}_{\mathsf{q}}[\pi(a \mid s)] 
%     % =  \argmax_{a \in \mathcal{A}} \sum_{\pi \in \Pi}\mathsf{q}(\pi)\pi(a \mid s)
%     = \argmax_{a \in \mathcal{A}} \sum_{\mu \in \mathsf{q}} \mathsf{q}(\mu) \mu(a | s).
% % \end{split}
% \end{equation*}

% We now formulate the imitation learning problem, of finding a policy $\pi$ that could explain expert demonstrations $\mathcal{D}$, as a Bayesian inference task, and present an efficient algorithm (Algorithm \ref{alg:inference}) for solving it approximately.

%We now describe an approximate Bayesian inference algorithm for imitation learning of policies in \texttt{LPP} from demonstrations $\mathcal{D}$. 
%We will define the policy prior $\mathsf{p}(\pi)$ and the likelihood $\mathsf{p}(\mathcal{D} \mid \pi)$ and then describe an algorithm for approximating the posterior $\mathsf{p}(\pi \mid \mathcal{D}) \propto \mathsf{p}(\mathcal{D} \mid \pi)\mathsf{p}(\pi)$.
%We can then use the approximate posterior over policies to derive a state-conditional posterior over actions, giving us a final stochastic policy $\pi_*$. 
% See Algorithm 1 in the appendix for pseudocode.
\subsection{Probabilistic Model $\mathsf{p}(\pi, \mathcal{D})$}
We begin by specifying a probabilistic model over policies and demonstrations $\mathsf{p}(\pi, \mathcal{D})$, which factors into a prior distribution $\mathsf{p}(\pi)$ over policies in $\texttt{LPP}$, and a likelihood $\mathsf{p}(\mathcal{D}\mid\pi)$ giving the probability that an expert generates demonstrations $\mathcal{D}$ by following the policy $\pi$.

We choose the prior distribution $\mathsf{p}(\pi)$ to encode a preference for those policies which use fewer, simpler feature detector programs. Recall that a policy $\pi \in \texttt{LPP}$ is parameterized by a logical formula $h(s, a) = \bigvee_{i=1\dots M} \left(\bigwedge_{j=1\dots N_i} f_{i, j}(s, a)^{b_{ij}} (1-f_{i,j}(s,a))^{1-b_{ij}}\right)$, in which each of the $f_{i, j}$ is a binary feature detector expressed in a simple DSL and the $b_{ij}$ are binary parameters that determine whether a given feature detector is negated. We set the prior probability of such a policy to depend only on the number and sizes of the programmatic components $f_{i, j}$: namely, $p(\pi) \propto \prod_{i=1}^{M}\prod_{j=1}^{N_i} \mathsf{p}(f_{i, j})$, the probability of generating each of the $f_{i, j}$ independently from a probabilistic context-free grammar $\mathsf{p}(f)$  \citep{manning1999foundations}.\footnote{Note that without some maximum limit on $\sum_{i=1}^MN_i$, this is an improper prior, and for this technical reason, we introduce a uniform prior on $\sum_{i=1}^M N_i$, between $1$ and a very high maximum value $\alpha$; the resulting factor of $\frac{1}{\alpha}$ does not depend on $\pi$ at all, and can be folded into the proportionality constant.} 
%Recall that a policy $\pi \in \texttt{LPP}$ is defined as a logical combination $h$ of many Boolean programs $f_{i, j}$, each of which is an expression in a DSL. We now place a prior, $\mathsf{p}(\pi)$, over policies, preferring those policies that are built from fewer and simpler programs.
%We factor the prior probability of a policy $\pi$ into a product of priors over its constituent programs: $\mathsf{p}(\pi) \propto \prod_{i=1}^M \prod_{j=1}^{N_i} \mathsf{p}(f_{i, j})$\footnote{Note that without some maximum limit on $\sum_{i=1}^MN_i$, this is an improper prior, and for this technical reason, we introduce a uniform prior on $\sum_{i=1}^M N_i$, between $1$ and a very high maximum value $\alpha$; the resulting factor of $\frac{1}{\alpha}$ does not depend on $\pi$ at all, and can be folded into the proportionality constant.}.
%If there were a finite number of expressible programs and $\mathsf{p}(f_{i, j})$ were uniform, this policy prior would favor policies with the fewest programs. 
%But we are instead in the regime where there are an infinite number of possible programs and some are much simpler than others. 
%We thus define a prior $\mathsf{p}$ over programs $f_{i,j}$ using a probabilistic grammar \citep{manning1999foundations}. 
%Given a probabilistic grammar, it is straightforward to enumerate programs with monotonically decreasing probability via best-first search. 
%We take advantage of this property during inference. 
The grammar we use in this work is shown in the appendix.

The likelihood of a dataset $\mathcal{D}$ given a policy $\pi$ is $\mathsf{p}(\mathcal{D} \mid \pi) \propto \prod_{i=1}^{N}\prod_{j=1}^{T_i} \pi(a_{ij} \mid s_{ij})$. 
In the appendix, we also consider the case where demonstrations are corrupted by noise.
%To derive a noise-tolerant likelihood model, we suppose that the expert policy is followed with probability $(1 - \epsilon)$, and with probability $\epsilon$, a random action is taken.
%(This is akin to the expert following an $\epsilon$-greedy policy.)
%The likelihood then becomes $\mathsf{p}(\mathcal{D} \mid \pi) \propto \prod_{i=1}^{N}\prod_{j=1}^{T_i} (1-\epsilon)\pi(a_{ij} \mid s_{ij}) + \frac{\epsilon}{|\mathcal{A}|}$.

\begin{table*}[h]
\centering
\begin{tabular}{ l l p{7cm}}
\multicolumn{1}{c}{\bf Method} &\multicolumn{1}{c}{\bf Type} &\multicolumn{1}{c}{\bf Description}
\\ \hline \\
$\texttt{cell\_is\_value}$ & V $\rightarrow$ C  & Check whether the attended cell has a given value \\
$\texttt{shifted}$  & O $\times$ C $\rightarrow$ C & Shift attention by an offset, then check a condition \\
$\texttt{scanning}$ & O $\times$ C $\times$ C $\rightarrow$ C  & Repeatedly shift attention by the given offset, and check which of two conditions is satisfied first \\
$\texttt{at\_action\_cell}$ & C $\rightarrow$ P & Attend to the action cell and check a condition \\
$\texttt{at\_cell\_with\_value}$ & V $\times$ C $\rightarrow$ P & Attend to a cell with the value and check condition  \\
\end{tabular}
\vspace{1mm}
\caption{Methods of the domain-specific language (DSL) used in this work. A \textit{program} (P) in the DSL implements a predicate on state-action pairs (i.e., P $ = \mathcal{S} \times \mathcal{A} \rightarrow \{0, 1\}$), by attending to a certain cell, then running a \textit{condition} (C). Conditions check that some property holds of the current state \textit{relative} to an implicit attention pointer. V ranges over possible grid cell values and an ``off-screen'' token, and O over ``offsets,'' which are pairs $(x, y)$ of integers specifying horizontal and vertical displacements.}
\label{table:dsl}
\end{table*}

\begin{figure*}[h]
    \centering
    \includegraphics[width=0.75\textwidth]{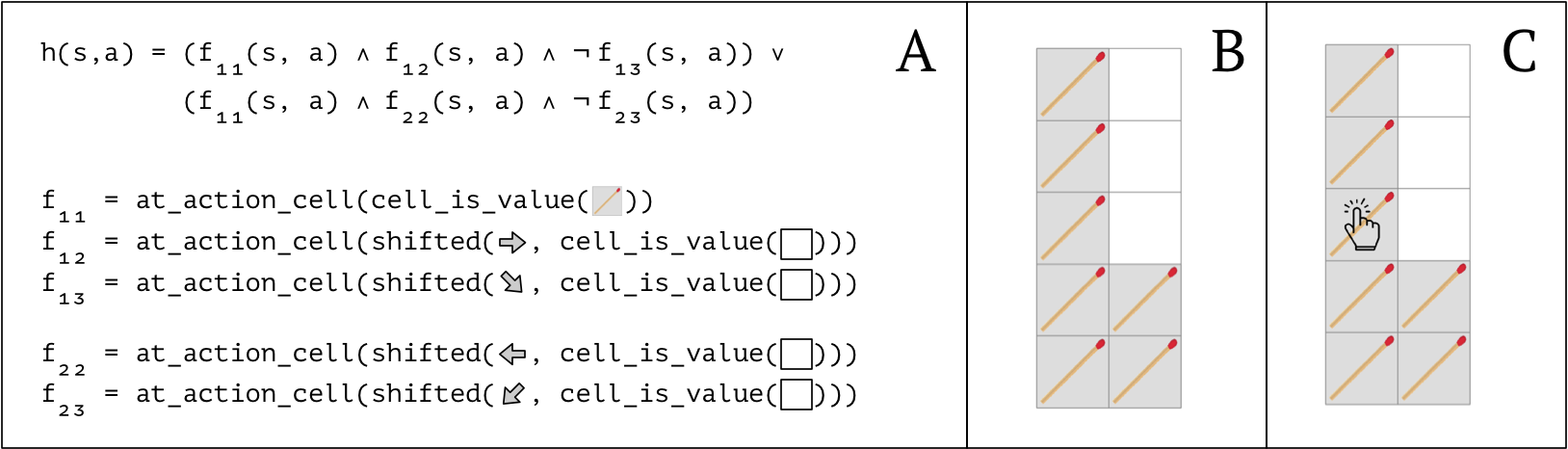}
    \caption{Example of a policy in \texttt{LPP} for the ``Nim'' game. (A) $h(s, a)$ is a logical combination of programs from a DSL. For example, $f_{12}$ returns True if the cell to the right of the action $a$ has value $\Box$. The induced policy is $\pi(a \mid s) \propto h(s, a)$. (B) Given state $s$, (C) there is one action selected by $h$. This policy encodes the ``leveling'' tactic, which wins the game.}
    \label{fig:nim_dsl_example}
\end{figure*}

\begin{figure*}[h]
    \centering
    \includegraphics[width=0.95\textwidth]{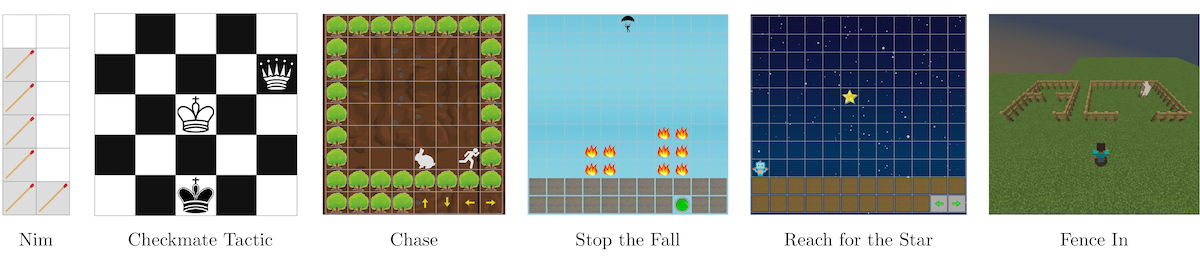}
    \caption{The strategy games studied in this work. See the appendix for descriptions and additional illustrations. }
    \label{fig:all_games}
\end{figure*}

\begin{figure*}[h]
    \centering
    \includegraphics[width=0.95\textwidth]{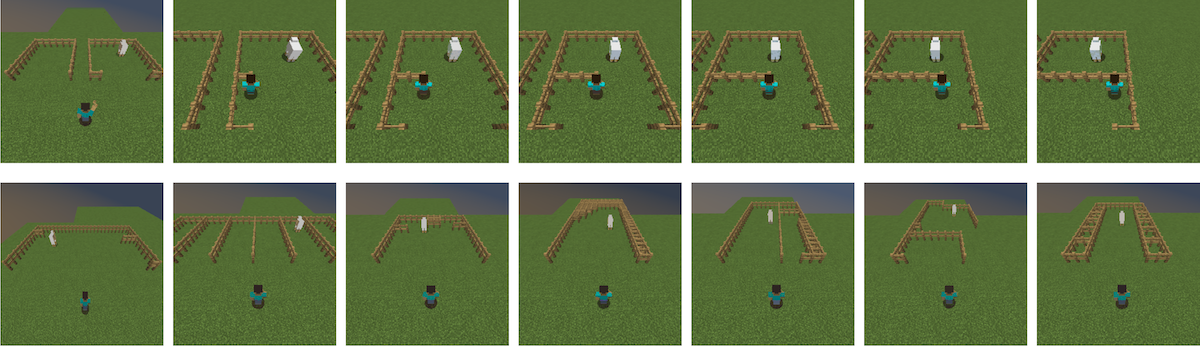}
    \caption{One-shot imitation learning in ``Fence In.'' From a single demonstration (top), we learn an enclosing strategy that generalizes to many new task instances (bottom).}
    \label{fig:fence_in_example}
\end{figure*}

\subsection{Approximating the Posterior $\mathsf{p}(\pi \mid \mathcal{D})$}

\begin{algorithm}[h]
\SetAlgoLined
\SetKwInOut{Input}{input}
 \Input{Demos $\mathcal{D}$, ensemble size $K$, max iters $L$}
Create anti-demos $\overline{\mathcal{D}} = \{ (s, a') : (s, a) \in \mathcal{D}, a' \neq a \}$\;
% Initialize features $X[(s, a)] = []$ for $(s, a) \in \mathcal{D} \cup \overline{\mathcal{D}}$\;
Set labels $y[(s, a)] = 1$ if $(s, a) \in \mathcal{D}$ else $0$\;
Initialize approximate posterior $\mathsf{q}$\;
  \For{$i$ in $1, ..., L$}{
  $f_i$ = generate\_next\_feature()\; %\tcp*{Search the grammar}
  $X = \{(f_1(s, a), ..., f_i(s, a))^T : (s, a) \in \mathcal{D} \cup \overline{\mathcal{D}}\}$
%   \For{$(s, a) \in \mathcal{D} \cup \overline{\mathcal{D}}$}{
%     $X[(s, a)]$.append($f_i(s, a)$)\;
%   }
  $\mu_i, w_i$ =  logical\_inference($X$, $y$, $\mathsf{p}(f)$, $K$)\;
  update\_posterior($\mathsf{q}$, $\mu_i$, $w_i$)\;
 }
 \Return{$\mathsf{q}$}\;
 \caption{\texttt{LPP} imitation learning}
 \label{alg:inference}
\end{algorithm}

We now have a prior $\mathsf{p}(\pi)$ and likelihood $\mathsf{p}(\mathcal{D} \mid \pi)$, and we wish to compute an approximate posterior $\mathsf{q}(\pi) \approx \mathsf{p}(\pi \mid \mathcal{D})$. 
%Computing this posterior exactly is intractable, so we instead aim to find a distribution $\mathsf{q}(\pi)$ that is close to the posterior. 
%Formally, our scheme is a variational inference algorithm, which iteratively minimizes the KL divergence from $\mathsf{q}$ to the true posterior. 
We take $\mathsf{q}$ to be a weighted mixture of $K$ policies $\mu_1, \dots, \mu_K$ (in our experiments, $K=25$) and initialize it so that each $\mu_i$ is equally weighted and equal to the uniform policy, $\mu_i(a \mid s) \propto 1$. Our core insight is a way to exploit the structure of \texttt{LPP} to efficiently search the space of policies and update the mixture $\mathsf{q}$ to better match the posterior. In the appendix, we show that this scheme is formally a variational inference algorithm that iteratively minimizes the KL divergence from $\mathsf{q}$ to the true posterior. 

Our algorithm is given a set of demonstrations $\mathcal{D}$. The state-action pairs $(s, a)$ in $\mathcal{D}$ comprise positive examples --- inputs for which $h(s, a) = 1$. We start by computing a set of ``anti-demonstrations''  $\overline{\mathcal{D}} = \{ (s, a') \mid (s, a) \in \mathcal{D}, a' \neq a \}$, which serve as approximate negative examples. ($\overline{\mathcal{D}}$ is approximate because it may contain false negatives, but they will generally constitute only a small fraction of the set.) 

We now have a binary classification problem with positive examples $\mathcal{D}$ and negative examples $\overline{\mathcal{D}}$. The main loop of our algorithm considers progressively larger feature representations of these examples. At iteration $i$, we use only the simplest feature detectors $f_1$, ..., $f_i$, where ``simplest'' here means ``of highest probability under the probabilistic grammar $\mathsf{p}(f)$.'' We can enumerate features in this order by performing a best-first search through the grammar.

Given a finite set of feature detectors $f_{1},\dots,f_{i}$, we can convert any state-action pair $(s, a)$ into a length-$i$ binary feature vector $\mathbf{x} \in \{0, 1\}^i = (f_1(s, a), \dots, f_i(s, a))^T$. We do this conversion on $\mathcal{D}$ and $\overline{\mathcal{D}}$ to obtain a design matrix $X_i \in \{0, 1\}^{|\mathcal{D} \cup \overline{\mathcal{D}}| \times i}$. The remaining problem of learning a binary classifier as a logical combination of binary features is very well understood \citep{mitchell1978version,valiant1985learning,quinlan1986induction,dietterich1986learning}. 
In this work, we use an off-the-shelf stochastic greedy decision-tree learner \cite{scikit-learn}.

Given a learned decision tree, we can easily read off a logical formula $h(s, a) = \bigvee_{j=1\dots M} \left(\bigwedge_{l=1\dots N_j} f_{j, l}(s, a)^{b_{jl}} (1-f_{j,l}(s,a))^{1-b_{jl}}\right)$, in which each of the $f_{j, l}$ is one of the $i$ feature detectors under consideration at iteration $i$. This induces a candidate policy $\mu_*(a | s) \propto h(s, a)$. We can evaluate its prior probability $\mathsf{p}(\mu_*)$ and its likelihood $\mathsf{p}(\mathcal{D}\mid\mu_*)$, then decide whether to include $\mu_*$ in our mixture $\mathsf{q}$, based on whether its unnormalized posterior probability is greater than that of the lowest-scoring existing mixture component. The mixture is always weighted according to our model over $\pi$ and $\mathcal{D}$, so that $\mathsf{q}(\mu_j) = \frac{\mathsf{p}(\mu_j \mid \mathcal{D})}{\sum_{i=1}^K \mathsf{p}(\mu_i \mid  \mathcal{D})}$. In practice, we run the decision-tree learner several times (5 in experiments) with different random seeds to generate several distinct candidate policies at each iteration of the algorithm. We can stop the process after a fixed number of iterations, or when the prior probabilities of the enumerated programs $f_i$ fall below a threshold: any policy that uses a feature detector $f_i$ with prior probability $\mathsf{p}(f_i) < \mathsf{p}(\mu_j, \mathcal{D})$ for all $\mu_j$ in $\mathsf{q}$'s support has no chance of meriting inclusion in our mixture.

Once we have an approximation $\mathsf{q}$ to the posterior, we can use it to derive a final policy for use at test time:
\begin{equation*}
% \begin{split}
    \pi_*(s) = \argmax_{a \in \mathcal{A}} \mathbb{E}_{\mathsf{q}}[\pi(a \mid s)] 
    % =  \argmax_{a \in \mathcal{A}} \sum_{\pi \in \Pi}\mathsf{q}(\pi)\pi(a \mid s)
    = \argmax_{a \in \mathcal{A}} \sum_{\mu \in \mathsf{q}} \mathsf{q}(\mu) \mu(a | s).
% \end{split}
\end{equation*}
We could alternatively use the full distribution over actions to guide exploration, e.g., in combination with reinforcement learning \citep{hester2018deep}. In this work, we focus on exploitation and therefore require only the maximum \textit{a posteriori} actions, for use with a deterministic final policy.

\section{Experiments and Results}

We now present experiments to evaluate the data efficiency, computational complexity, and generalization of \texttt{LPP} versus several baselines. We also analyze the learned policies, examine qualitative performance (see Figure \ref{fig:fence_in_example} and the appendix), and conduct ablation studies to measure the contributions of the components of \texttt{LPP}. All experiments were performed on a single laptop running macOS Mojave with a 2.9 GHz Intel Core i9 processor and 32 GB of memory.

% We now study the extent to which policies in \texttt{LPP} can be learned from a few demonstrations. For each of five tasks, we evaluate \texttt{LPP} as a function of the number of demonstrations in the training set and the number of programs enumerated. We examine the performance of the learned policies empirically and qualitatively and analyze the complexity of the discovered programs. We further compare the performance of the learned policies against several baselines including two different types of convolutional networks, local linear policies, and program-based policies learned via enumeration. We then perform ablation studies to dissect the contributions of the main components of \texttt{LPP}: the programmatic feature detectors, the decision tree learning, and the probabilistic grammar prior.

\subsection{Tasks}

We consider six diverse strategy games (Figure \ref{fig:all_games}) that share a common state space ($\mathcal{S} = \bigcup_{h,w \in \mathbb{N}} V^{hw}$; variable-sized grids with discrete-valued cells) and action space $\mathcal{A} = \mathbb{N} \times \mathbb{N}$; single ``clicks'' on any cell). For a grid of dimension $h \times w$, we only consider clicks on the grid, i.e., $\{1, ..., h \} \times  \{1, ..., w \}$. Grid sizes vary within tasks. These tasks feature high variability between different task instances; learning a robust policy requires substantial generalization. The tasks are also very challenging due to the unbounded action space, the absence of shaping or auxiliary rewards, and the arbitrarily long horizons that may be required to solve a task instance.
Each task has a maximum episode length of 60 and counts as a failure if the episode terminates without a success. There are 11 training and 9 test instances per task. Instances of Nim, Checkmate Tactic, and Reach for the Star are procedurally generated; instances of Stop the Fall, Chase, and Fence In are manually generated, as the variation between instances is not trivially parameterizable. We provide descriptions of the tasks in the appendix.
%and encourage the reader to see the appendixary video for examples.

\subsection{Domain-Specific Language}

Recall that each feature detector program takes a state and action as input and returns a Boolean value. 
In our tasks, states are full grid layouts and actions are single grid cells (``clicks'').
The specific DSL of feature detectors that we use in this work (Table \ref{table:dsl}) is inspired by early work in visual routines \citep{ullman1987visual,hay2018behavior}. 
Each program implements a procedure for attending to some grid cell and checking that a local condition holds nearby. 
Given input $(s, a)$, a program begins by initializing an implicit attention pointer either to the grid cell in $s$ associated with action $a$ (\texttt{at\_action\_cell}), or to an arbitrary grid cell containing a certain value (\texttt{at\_cell\_with\_value}). 
Next, the program will check a condition at or near the attended cell. 
The simplest condition is \texttt{cell\_is\_value}, which checks whether the attended cell has a certain value. 
More complex conditions, which look not just \textit{at} but \textit{near} the attended cell, can be built up using the \texttt{shifted} and \texttt{scanning} methods. 
The \texttt{shifted} method builds a condition that first shifts the attention pointer by some offset, then applies another condition. 
The \texttt{scanning} method starts at the currently attended cell and ``scans'' along some direction, repeatedly shifting the attention pointer by a specified offset and checking whether either of two specified conditions hold. 
If, while scanning, the first condition becomes satisfied before the second, the \texttt{scanning} condition returns $1$. 
Otherwise, it returns $0$. 
Thus the overall DSL contains five methods, which are summarized in Table \ref{table:dsl}.
See Figure \ref{fig:nim_dsl_example} for a complete example of a policy in \texttt{LPP} using this DSL.

\subsection{Baselines}

% We compare \texttt{LPP} against four baselines: 

%two from deep learning, one using local features alone, and one from program induction. We selected these baselines to encode varying degrees of domain-specific structure. For the deep and linear baselines, we pad the inputs so that all states are the same maximal width and height.

%(If larger states were added to the training or test sets, these baselines would require retraining.) 

\textbf{Local Linear Network (LLN)}: A single $3 \times 3$ convolutional filter is trained to classify whether each cell in $s$ should be ``clicked,'' based only on the 8 surrounding cells.
\textbf{FCN}: A deep fully convolutional network \citep{long2015fully} is trained with the same inputs and outputs as ``Local Linear.'' The network has 8 convolutional layers with kernel size 3, stride 1, padding 1, 4 channels (8 in the input layer), and ReLU nonlinearities. This architecture was chosen to reflect the receptive field sizes we expect are necessary for the tasks.
\textbf{CNN}: A standard convolutional neural network is trained with full grid inputs and discrete action outputs. Grids are padded so that all have the same maximal height and width. The architecture is: 64-channel convolution; max pooling; 64-channel fully-connected layer; $|\mathcal{A}|$-channel fully-connected layer. All kernels have size 3 and all strides and paddings are 1.
\textbf{Vanilla Program Induction (VPI)}: Full policies are enumerated from a DSL grammar that includes logical disjunctions, conjunctions, and negations over the feature detector DSL. The number of disjunctions and conjunctions each follow a geometric distribution ($p=0.5$). (Several other values of $p$ were also tried without improvement.) Policies are then enumerated and mixed as in \texttt{LPP} learning; this baseline is thus identical to \texttt{LPP} learning but with the greedy Boolean learning removed.

% The fourth baseline (``Enumeration'') is full policy enumeration. The original DSL grammar  is extended to include logical disjunctions, conjunctions, and negations over constituent programs so that full policies are enumerated in DNF form. The number of disjunctions and conjunctions each follow a geometric distribution ($p=0.5$). (Several other values of $p$ were also tried without improvement.) Policies are then enumerated and mixed as in \texttt{LPP} learning; this baseline is thus identical to \texttt{LPP} learning but with the greedy Boolean learning removed.

% \begin{figure*}[t]
%     \centering
%     \includegraphics[width=1.0\textwidth]{fence_in_example.png}
%     \caption{One-shot imitation learning in ``Fence In.'' From a single demonstration (top), we learn an enclosing strategy that generalizes to many new task instances (bottom).}
%     \label{fig:fence_in_example}
% \end{figure*}

% \begin{figure}[t]
%     \centering
%     \includegraphics[width=1.0\textwidth]{stf_learned_clause.png}
%     \caption{One of the clauses of the learned MAP policy for ``Stop the Fall''. The clause suggests clicking a cell if scanning up we find a parachuter before a drawn or static block; if the cell is empty; if the cell above is not drawn; if the cell is not a green button; and if a static block is immediately below. Note that this clause is slightly redundant and may trigger an unnecessary action for a task instance where there are no ``fire'' cells.}
%     \label{fig:stf_learned_clause}
% \end{figure}

%\subsection{Learning and Generalizing from Demonstrations}

\subsection{Effect of Number of Demonstrations}

\begin{figure*}[t]
    \centering
    \includegraphics[width=0.98\textwidth]{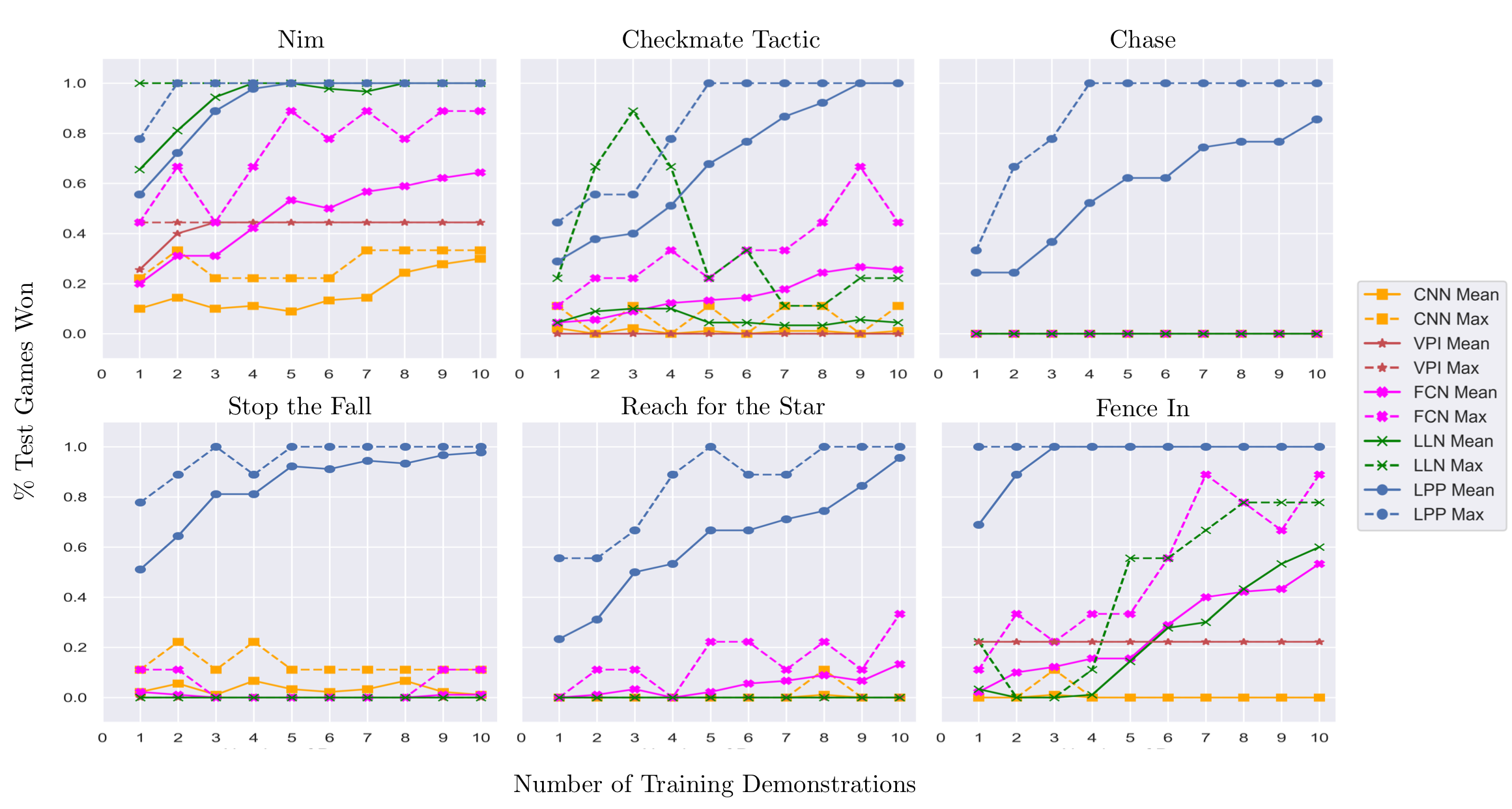}
    \caption{Performance on held-out test task instances as a function of the number of training demonstrations for \texttt{LPP} (ours) and four baselines. Maximums and means are over 10 training sets.}
    \label{fig:num_demos_results}
\end{figure*}

We first evaluate the test-time performance of \texttt{LPP} and baselines as the number of training demonstrations varies from 1 to 10.  For each number of demonstrations, we run leave-one-out cross validation: 10 trials, each featuring a distinct set of demonstrations drawn from the overall pool of 11 training demonstrations. \texttt{LPP} learning is run for 10,000 iterations for each task. The mean and maximum trial performance offer complementary insight: the mean reflects the expected performance if demonstrations were selected at random; the maximum reflects the expected performance if the most useful demonstrations were selected, perhaps by an expert teacher.
Results are shown in Figure \ref{fig:num_demos_results}. On the whole, \texttt{LPP} markedly outperforms all baselines, especially on the more difficult tasks. The baselines are limited for different reasons. The highly parameterized CNN baseline is able to perfectly fit the training data and win all \textit{training} games (not shown), but given the limited training data and high variation from training to task, it severely overfits and fails to generalize. The FCN baseline is also able to fit the training data almost perfectly. Its additional structure permits better generalization in Nim, Checkmate Tactic, Reach for the Star, and Fence In than the CNN, but overall its performance is still far behind \texttt{LPP}. In contrast, the LLN baseline is unable to fit the training data; with the exception of Nim, its training performance is close to zero. Similarly, the training performance of the VPI baseline is near or at zero for all tasks beyond Nim. In Nim, there is evidently a low complexity program that works roughly half the time, but an optimal policy is more difficult to find.

\subsection{Effect of Number of Programs Searched}

\begin{figure}[t]
    \centering
    \includegraphics[width=0.48\textwidth]{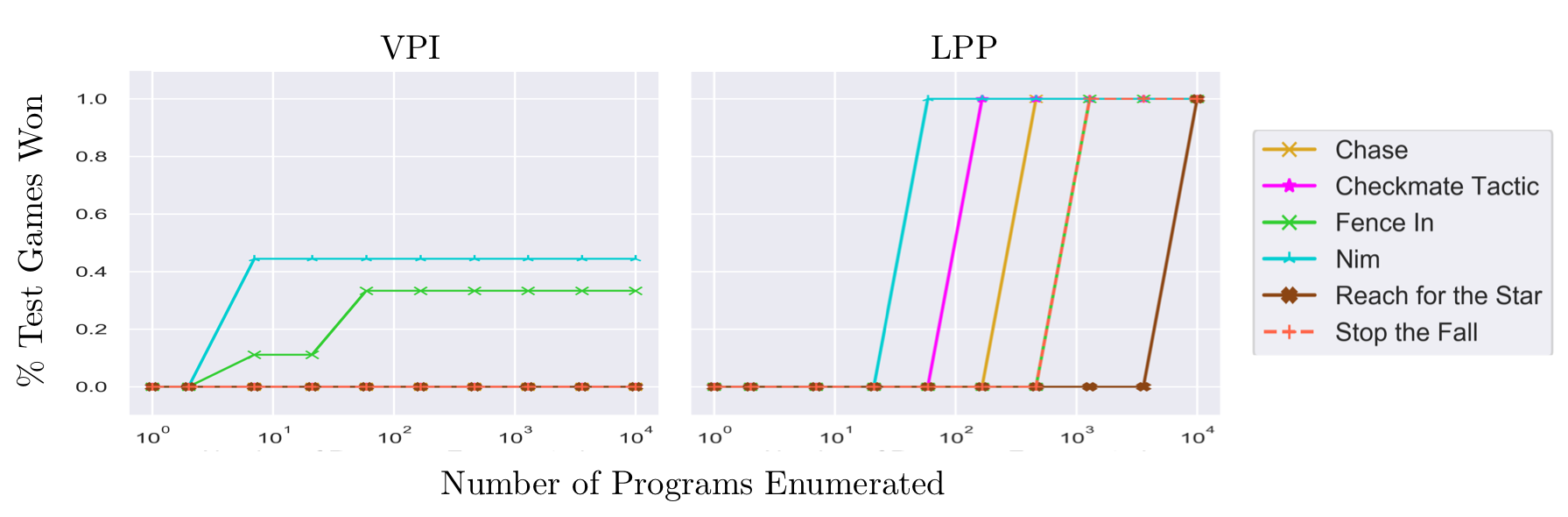}
    \caption{Performance on held-out test task instances as a function of the number of programs enumerated for the Vanilla Program Induction (VPI) baseline and \texttt{LPP} (ours).}
    \label{fig:num_programs_results}
\end{figure}

\begin{figure}[h]
    \centering
    \includegraphics[width=0.48\textwidth]{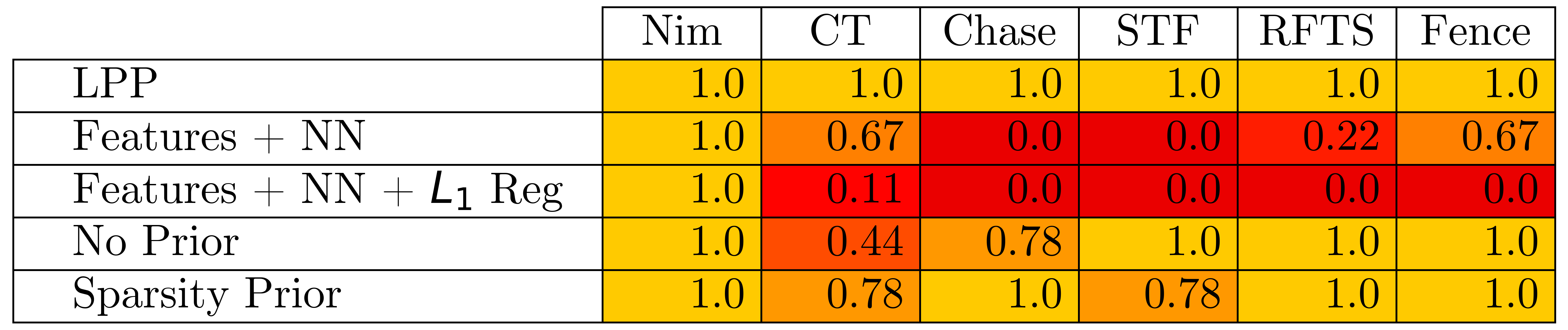}
    \caption{Performance on held-out test task instances for \texttt{LPP} and four ablation models.}
    \label{fig:ablation_table}
\end{figure}

We now examine test-time performance of \texttt{LPP} and VPI as a function of the number of programs searched. For this experiment, we give both methods all 11 training demonstrations for each task. Results are shown in Figure \ref{fig:num_programs_results}. \texttt{LPP} requires fewer than 100 programs to learn a winning policy for Nim, fewer than 1,000 for Checkmate Tactic and Chase, and fewer than 10,000 for Stop the Fall, Reach for the Star, and Fence In. In contrast, VPI is unable to achieve nonzero performance for any task other than Nim, for which it achieves roughly 45\% performance after 100 programs enumerated. 
The lackluster performance of VPI is unsurprising given the combinatorial explosion of programs. For example, the optimal policy for Nim shown in Figure \ref{fig:nim_dsl_example} involves six constituent programs, each with a parse tree depth of three or four. There are 108 unique constituent programs with parse tree depth three and therefore more than 13,506,156,000 full policies with six or fewer constituent programs. VPI would have to search roughly so many programs before arriving at a winning policy for Nim, which is by far the simplest task. In contrast, a winning \texttt{LPP} policy is learnable after fewer than 100 enumerations. In practical terms, \texttt{LPP} learning for Nim takes on the order of 1 second on a laptop without highly optimized code; after running VPI for six hours in the same setup, a winning policy is still not found.

\subsection{Ablation Studies}

We now perform ablation studies to explore which aspects of the \texttt{LPP} class and learning algorithm contribute to the strong performance. We consider four ablated models. The ``Features + NN'' model learns a neural network state-action binary classifier on the first 10,000 feature detectors enumerated from the DSL. This model addresses the possibility that the features alone are powerful enough to solve the task when combined with a simple classifier. The NN is a multilayer perceptron with two layers of dimension 100 and ReLU activations. The ``Features + NN + $L_1$ Regularization'' model is identical to the previous baseline except that an $L_1$ regularization term is added to the loss to encourage sparsity. This model addresses the possibility that the features alone suffice when we incorporate an Occam's razor bias similar to the one that exists in \texttt{LPP} learning. The ``No Prior'' model is identical to \texttt{LPP} learning, except that the grammatical prior is replaced with a uniform prior. Similarly, the ``Sparsity Prior'' model uses a prior that penalizes the number of top-level programs involved in the policy, without regard for the relative priors of the individual programs. Results are presented in Figure \ref{fig:ablation_table}. They confirm that the each component --- the feature detectors, the sparsity regularization, and the grammatical prior --- adds value to the overall framework.

\section{Related Work}
\label{sec:related-work}
Sample efficiency and generalization are two of the main concerns in imitation learning \cite{schaal1997learning,abbeel2004apprenticeship}. 
To cope with limited data, %e.g., due to robotic resource constraints \citep{billard2008robot, argall2009survey, konidaris2012robot},
demonstrations can be used in combination with additional RL \cite{hester2018deep,nair2018overcoming}. Alternatively, a mapping from demonstrations to policies can be learned from a background set of tasks \citep{duan2017one,finn2017one}. A third option is to introduce a prior over a structured policy class \citep{andre2002state,NIPS2010_3992,wingate2011bayesian}, e.g., hierarchical policies \citep{niekum2013semantically}. Our work fits into the third tradition; our contribution is a new policy class with a structured prior that enables efficient learning.

\texttt{LPP} policies are logical at the ``top level'' and programmatic at the ``bottom level.'' Logical representations for RL problems have been considered in many previous works, particularly in \textit{relational RL} 
\citep{dvzeroski1998relational,natarajan2011imitation}. Also notable is work by \citet{shah2018bayesian}, who learn linear temporal logic specifications from demonstration using a finite grammatical prior. While some \texttt{LPP} programs may be seen as fixed-arity logical relations in the classical sense, others are importantly more general and powerful, involving loops and a potentially arbitrary number of atoms. For example, one program suffices to check whether a Queen's diagonal path is clear in Chess; no relation over a fixed number of squares can capture the same feature. Furthermore, relational RL assumes that relations are fixed, finite, and given, typically hand-designed by the programmer. In \texttt{LPP}, the programmer instead supplies a DSL describing infinitely many features.

\texttt{LPP} learning is a particular type of \textit{program synthesis}, which more broadly refers to a search over programs, including but not limited to the case where we have a grammar over programs, the programs are mappings, and input-output examples are available. 
When a grammar is given, the problem is sometimes called \textit{syntax-guided synthesis} \citep{alur2013syntax}. 
Most relevant is work by \citet{alur2017scaling}, who propose a ``divide and conquer'' approach that uses greedy decision tree learning in combination with enumeration from a grammar of ``conditions'', similar to our ``No Prior'' baseline. 

Recent work has also examined \textit{neural program synthesis} (NPS) wherein a large dataset of (input, output, program) examples is used to train a guidance function for program enumeration \citep{parisotto2016neuro,devlin2017robustfill,bunel2018leveraging,huang2019neural}.
%The enumeration is often more efficient than VPI, since the search through the grammar for a program is directly influenced by the given input-output example. 
In practice, NPS methods are still limited to programs involving $\sim$10 primitives. 
The neural guidance can delay, but not completely avoid, the combinatorial explosion of search in program space. 
\texttt{LPP} learning is not a generic program induction method, but rather, an algorithm that exploits the logical structure of \texttt{LPP} programs to dramatically speed up search, sometimes finding programs with $\sim$250 primitives (Appendix Table 2). 
%In principle, neural guidance could be combined with \texttt{LPP} learning to push performance even further.

The interpretation of policy learning as an instance of program synthesis is explored in prior work \citep{wingate2011bayesian,sun2018neural,verma2018programmatically}. 
In particular, \citet{lazaro2018beyond} learn object manipulation concepts from before/after image pairs that can be transferred between 2D simulation and a real robot. 
In this work, we focus on the problem of \emph{efficient inference} and compare against vanilla program induction in experiments. 
%and make use of full demonstrations of a policy (instead of start/end configurations alone). 

% We define policies in terms of programs \citep{wingate2013compositional,xu2018neural,verma2018programmatically} and a prior over policies using a probabilistic grammar \citep{goodman2008rational,goodman2014concepts,piantadosi2016logical}. 
% Similar priors appear in Bayesian concept learning from cognitive science \citep{tenenbaum1999bayesian,tenenbaum2001generalization,lake2015human,ellis2018learning2,ellis2018learning}.
% %Incorporating insights from program induction for planning has seen renewed interest in the past year. 
% Of particular note is the work by \citet{lazaro2018beyond}, who learn object manipulation concepts from before/after image pairs that can be transferred between 2D simulation and a real robot. 
% %They define a DSL that involves visual perception with shifting attention, working memory, new object imagination, and object manipulation.
% %We use a DSL that similarly involves shifting attention and object-based (grid cell-based) actions.
% In this work, we focus on the problem of \emph{efficient inference} and compare against vanilla program induction in experiments. 
% %and make use of full demonstrations of a policy (instead of start/end configurations alone). 

\section{Discussion and Conclusion}

In an effort to efficiently learn policies from very few demonstrations that generalize substantially, we have introduced the \texttt{LPP} policy class and an approximate Bayesian inference algorithm for imitation learning. We have seen that the \texttt{LPP} policy class includes winning policies for a diverse set of strategy games, and moreover, that those policies can be efficiently learned from five or fewer demonstrations. 
%We have also confirmed that it would be intractable to learn the same policies with vanilla program induction, and that deep convolutional policies are prone to severe overfitting in this low data regime. In ablation studies, we found that each of the main components of the \texttt{LPP} representation and the learning algorithm contribute substantively to the overall strong performance.
In ongoing work we are studying how to scale our approach to a wider range of tasks, starting with more sophisticated DSLs that include counting or simple data structures. However, even our current DSL is surprisingly general.  For instance, in preliminary experiments (see appendix), we find that our current algorithm can learn a generalizing policy for Atari Breakout from just one demonstration.

Beyond policy learning, this work contributes to the long and ongoing discussion about the role of prior knowledge in AI. 
In the common historical narrative, early attempts to incorporate prior knowledge via feature engineering failed to scale, leading to the modern shift towards domain-agnostic deep learning methods \citep{sutton_2019}. 
Now there is renewed interest in incorporating inductive bias into contemporary methods, especially for problems where data is scarce. 
We argue that encoding prior knowledge via a probabilistic grammar over feature detectors and learning to combine these feature detectors with Boolean logic is a promising path forward. 
More generally, we submit that ``meta-feature engineering'' of the sort exemplified here strikes an appropriate balance between the strong inductive bias of classical AI and the flexibility and scalability of modern methods.

\section{Acknowledgements}
We gratefully acknowledge support from NSF grants 1523767 and 1723381; from ONR grant N00014-13-1-0333; from AFOSR grant   FA9550-17-1-0165;   from   ONR   grant   N00014-18-1-2847; from Honda Research; and from the Center for Brains, Minds  and  Machines  (CBMM),  funded  by  NSF  STC  award CCF-1231216. KA acknowledges support from NSERC. Any opinions,  findings,  and  conclusions  or  recommendations  expressed  in  this  material  are  those  of  the  authors  and  do  not necessarily reflect the views of our sponsors.

\bibliography{main.bib}
\bibliographystyle{aaai}

\appendix

\section{Probabilistic Grammar for DSL}

\renewcommand{\arraystretch}{1.3}
\begin{table}[h]
\begin{center}
\begin{tabular}{ | l | l | }
\multicolumn{1}{c}{Production rule} & \multicolumn{1}{c}{Probability} \\
\hline
\textbf{Programs} & \\
P $\rightarrow \texttt{at\_cell\_with\_value(}$V, C$\texttt{)}$ & $0.5$ \\
P $\rightarrow \texttt{at\_action\_cell(}$C$\texttt{)}$ & $0.5$ \\
\hline
\textbf{Conditions} & \\
C $\rightarrow \texttt{shifted(}$O, B$\texttt{)}$ & $0.5$ \\
C $\rightarrow$ B & $0.5$ \\
\hline
\textbf{Base conditions} & \\
B $\rightarrow \texttt{cell\_is\_value(}$V$\texttt{)}$ & $0.5$ \\
B $\rightarrow \texttt{scanning(}$O, C, C$\texttt{)}$ & $0.5$ \\
\hline
\textbf{Offsets} & \\
O $\rightarrow ($N, $0)$ & $0.25$ \\
O $\rightarrow (0$, N$)$ & $0.25$ \\
O $\rightarrow ($N, N$)$ & $0.5$ \\
\hline
\textbf{Numbers} & \\
N $\rightarrow \mathbb{N}$ & $0.5$ \\
N $\rightarrow -\mathbb{N}$ & $0.5$ \\
\hline 
\textbf{Natural numbers} (for $i = 1, 2, \dots$) & \\
$\mathbb{N} \rightarrow i$ & $(0.99)(0.01)^{i-1}$ \\
\hline
\textbf{Values} (for each value $\texttt{v}$ in this game) & \\
V $\rightarrow \texttt{v}$ & $1/|$V$|$ \\
\hline
\end{tabular}
\vspace{5mm}
\caption{The prior $\mathsf{p}(f)$ over programs, specified as a probabilistic context-free grammar (PCFG). }
\label{table:pcfg}
\end{center}
\end{table}

\section{Variational Interpretation of \texttt{LPP} Learning}

The algorithm for \texttt{LPP} learning presented in the main text can be understood as performing variational inference, iteratively minimizing the KL divergence from $\mathsf{q}$ to the true posterior. We make two observations to this effect. First, fixing the $K$ component policies in the support of the mixture $\mathsf{q}$, the component weights that minimize the KL divergence $D_{KL}(\mathsf{q}(\pi) \mid\mid \mathsf{p}(\pi \mid \mathcal{D}))$ are $\mathsf{q}(\mu_j) = \frac{\mathsf{p}(\mu_j \mid  \mathcal{D})}{\sum_{i=1}^K \mathsf{p}(\mu_i \mid \mathcal{D})}$, that is, each policy should be weighted according to its relative posterior probability. Second, if some policy $\mu_*$ not in the support of $\mathsf{q}$ has higher posterior probability than some existing mixture component $\mu_i$, a lower KL is always achievable by replacing $\mu_i$ with $\mu_*$ and reweighting the mixture accordingly. Our algorithm maintains a list $\mu_1, \dots, \mu_K$ of the $K$ best policies seen so far, and iteratively searches an increasingly large space for new policies $\mu_*$ of high posterior probability. Whenever a $\mu_*$ is found that has higher (unnormalized) posterior probability than the lowest-scoring policy $\mu_i$ in our ``best-$K$'' list, replace $\mu_i$ with $\mu_*$. At each iteration, our variational approximation is simply $\mathsf{q}(\mu_j) = \frac{\mathsf{p}(\mu_j \mid \mathcal{D})}{\sum_{i=1}^K \mathsf{p}(\mu_i \mid \mathcal{D})}$ for $j \in \{1, \dots, K\}$.

\section{Environment Details}

\subsection{Nim}

\subsubsection{Task Description}There are two columns (piles) of matchsticks and empty cells. Clicking on a matchstick cell changes all cells above and including the clicked cell to empty; clicking on a empty cell has no effect. After each matchstick cell click, a second player takes a turn, selecting another matchstick cell. The second player is modeled as part of the environment transition and plays optimally. When there are multiple optimal moves, one is selected randomly. The objective is to remove the \textit{last} matchstick cell.

\subsubsection{Task Instance Distribution} Instances are generated procedurally. The height of the grid is selected randomly between 2 and 20. Th    e initial number of matchsticks in each column is selected randomly between 1 and the height with the constraint that the two columns cannot be equal. All other grid cells are empty.

\subsubsection{Expert Policy Description} The winning tactic is to ``level'' the columns by selecting the matchstick cell next to a empty cell and diagonally up from another matchstick cell. Winning the game requires perfect play.

\subsection{Checkmate Tactic}

\subsubsection{Task Description} This task is inspired by a common checkmating pattern in Chess. Note that only three pieces are involved in this game (two kings and a white queen) and that the board size may be $H \times W$ for any $H, W$, rather than the standard $8 \times 8$. Initial states in this game feature the black king somewhere on the boundary of the board, the white king two cells adjacent in the direction away from the boundary, and the white queen attacking the cell in between the two kings. Clicking on a white piece (queen or king) \textit{selects} that piece for movement on the next action. Note that a selected piece is a distinct value from a non-selected piece. Subsequently clicking on an empty cell moves the selected piece to that cell if that move is legal. All other actions have no effect. If the action results in a checkmate, the game is over and won; otherwise, the black king makes a random legal move.

\subsubsection{Task Instance Distribution} Instances are generated procedurally. The height and width are randomly selected between 5 and 20. A column for the two kings is randomly selected among those not adjacent to the left or right sides. A position for the queen is randomly selected among all those spaces for which the queen is threatening checkmate between the kings. The board is randomly rotated among the four possible orientations.

\subsubsection{Expert Policy Description}  The winning tactic selects the white queen and moves it to the cell between the kings.

\subsection{Chase}

\subsubsection{Task Description}This task features a stick figure agent, a rabbit adversary, walls, and four arrow keys. At each time step, the adversary randomly chooses a move (up, down, left, or right) that increases its distance from the agent. Clicking an arrow key moves the agent in the corresponding direction. Clicking a gray wall has no effect other than advancing time. Clicking an empty cell creates a new (blue) wall. The agent and adversary cannot move through gray or blue walls. The objective is to ``catch'' the adversary, that is, move the agent into the same cell. It is not possible to catch the adversary without creating a new wall; the adversary will always be able to move away before capture.

\subsubsection{Task Instance Distribution} There is not a trivial parameterization to procedurally generate these task instances, so they are manually generated.

\subsubsection{Expert Policy Description}  The winning tactic advances time until the adversary reaches a corner, then builds a new wall next to the adversary so that it is trapped on three sides, then moves the agent to the adversary.

\begin{figure*}[t]
    \centering
    \includegraphics[width=0.8\textwidth]{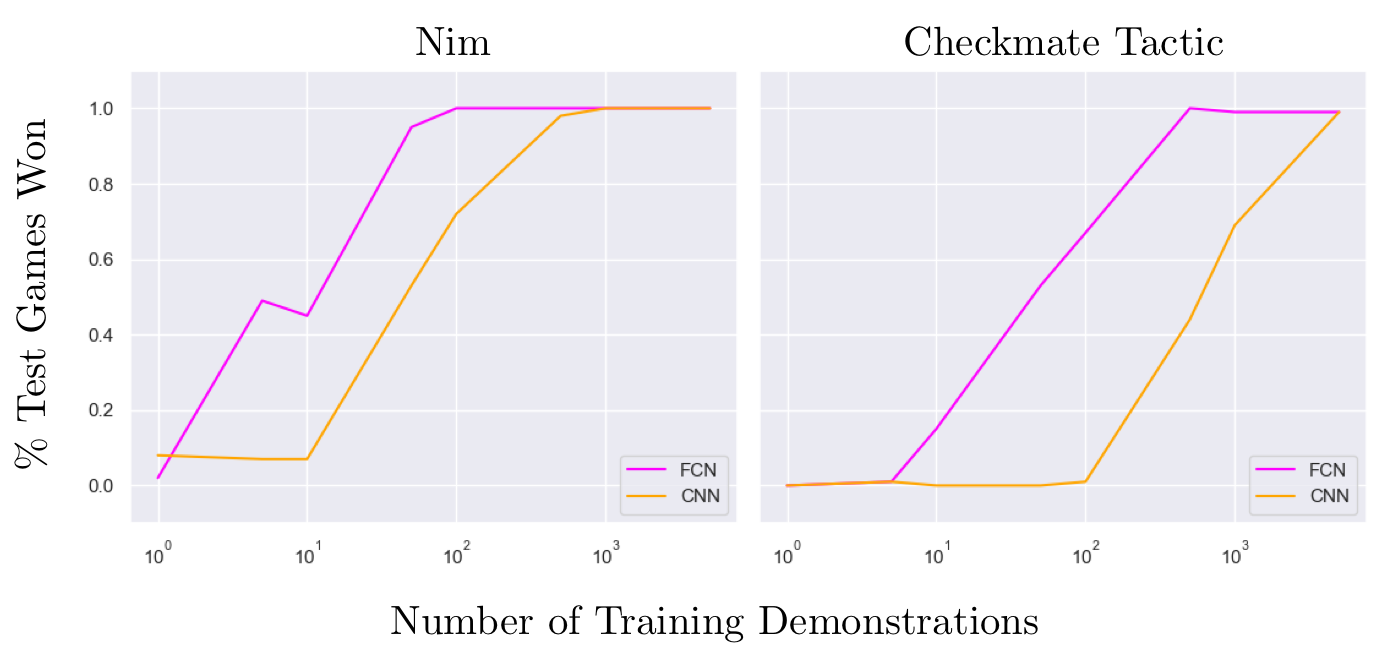}
    \caption{Performance on held-out test task instances as a function of the number of training demonstrations for CNN and FCN baselines.}
    \label{fig:data_effiency}
\end{figure*}

\subsection{Stop the Fall}

\subsubsection{Task Description}This task involves a parachuter, gray static blocks, red ``fire'', and a green button that turns on gravity and causes the parachuter and blocks to fall. Clicking an empty cell creates a blue static block. The game is won when gravity is turned on and the parachuter falls to rest without touching (being immediately adjacent to) fire.

\subsubsection{Task Instance Distribution} There is not a trivial parameterization to procedurally generate these task instances, so they are manually generated.

\subsubsection{Expert Policy Description} The winning tactic requires building a stack of blue blocks below the parachuter that is high enough to prevent contact with fire, and then clicking the green button.

\subsection{Reach for the Star}

\subsubsection{Task Description}In this task, a robot must move to a cell with a yellow star. Left and right arrow keys cause the robot to move. Clicking on an empty cell creates a dynamic brown block. Gravity is always on, so brown objects fall until they are supported by another brown block. If the robot is adjacent to a brown block and the cell above the brown block is empty, the robot will move on top of the brown block when the corresponding arrow key is clicked. (In other words, the robot can only climb one block, not two or more.) 

\subsubsection{Task Instance Distribution} Instances are generated procedurally. A height for the star is randomly selected between 2 and 11 above the initial robot position. A column for the star is also randomly selected so that it is between 0 and 5 spaces from the right border. Between 0 and 5 padding rows are added above the star. The robot position is selected between 0 and 5 spaces away from where the start of the minimal stairs would be. Between 0 and 5 padding columns are added to the left of the agent. The grid is flipped along the vertical axis with probability 0.5.

\subsubsection{Expert Policy Description} The winning tactic requires building stairs between the star and robot and then moving the robot up them.

\subsection{Fence In}

\subsubsection{Task Description} This task uses the Malmo interface to Minecraft \cite{johnson2016malmo}. As in the other tasks, observations are 2D grids with discrete values and actions are ``clicks'' on the grid. This task features fences, one sheep, and open space in the grid. Clicking on a grid cell has the effect of building a fence at that location if it is currently open. The objective of the task is to build fences in such a way that the sheep is completely enclosed by fences. (Enclosure is calculated by finding the connected component that contains the sheep and checking whether it reaches the boundary of the grid.) 

\subsubsection{Task Instance Distribution} There is not a trivial parameterization to procedurally generate these task instances, so they are manually generated.

\subsubsection{Expert Policy Description} The expert policy builds fences from left to right one cell below the sheep, starting at the nearest fence on the left and finishing at the nearest fence on the right.

\section{Additional Results}

\subsection{Data Efficiency of Baselines}

We saw in our main experiments that generalizable convolutional and fully convolutional policies cannot be learned in our regime of very low data and very high variation between task instances. In an effort to better understand the gap in data efficiency between the deep policies (CNN and FCN) and \texttt{LPP}, we again measure the test performance of the baselines as a function of the number of demonstrations, but with many more demonstrations than we previously considered. In the main paper, we report results for up to 10 demonstrations; here we report results for up to 5,000. 
We focus on Nim and Checkmate Tactic for this experiment.
%We use Nim, Checkmate Tactic, and Reach for the Star for this experiment, as these three tasks are procedurally generated. 
Recall that at most 5 demonstrations were necessary to learn a winning (test performance 1.0) policy in \texttt{LPP}. In Figure \ref{fig:data_effiency}, we see that FCNs require 100 demonstrations in Nim and 500 demonstrations in Checkmate Tactic to achieve test performance 1.0. We also find that CNNs require 1,000 demonstrations in Nim and 5,000 demonstrations in Checkmate Tactic. Thus \texttt{LPP} achieves 20 -- 1,000x data efficiency gains over these baselines.

\begin{figure*}[h]
    \centering
    \includegraphics[width=0.8\textwidth]{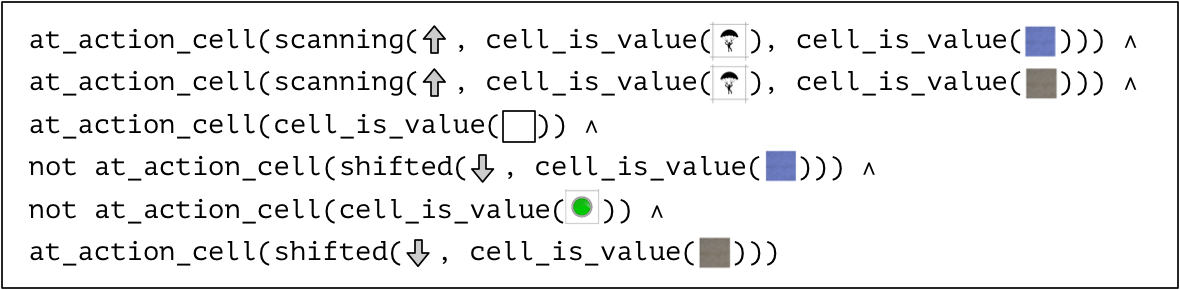}
    \caption{One of the clauses of the learned MAP policy for ``Stop the Fall''. The clause suggests clicking a cell if scanning up we find a parachuter before a drawn or static block; if the cell is empty; if the cell above is not drawn; if the cell is not a green button; and if a static block is immediately below. Note that this clause is slightly redundant and may trigger an unnecessary action for a task instance where there are no ``fire'' cells.}
    \label{fig:stf_learned_clause}
\end{figure*}
\begin{table*}[h]
\begin{center}
\begin{tabular}{lllll}
\multicolumn{1}{c}{\bf Task}  &\multicolumn{1}{c}{\bf Number of Programs} &\multicolumn{1}{c}{\bf Method Calls}  &\multicolumn{1}{c}{\bf Max Depth}
\\ \hline
Nim         & 11 & 32 & 4  \\
Checkmate Tactic         & 23 & 60 & 3  \\
Chase         &  97 & 244 & 3  \\
Stop the Fall         & 17 & 50 & 4  \\
Reach for the Star         & 34 & 151 & 4  \\
Fence In         & 9 & 40 & 3  \\

\end{tabular}
\vspace{1mm}
\caption{Analysis of the MAP policies learned in \texttt{LPP} with 10,000 programs enumerated and all 11 training demonstrations. The number of top-level programs, the number of constituent method calls, and the maximum parse depth of a program in the policy is reported for each of the five tasks.}
\label{table:mapprograms}
\end{center}
\end{table*}

\subsection{Learned Policy Analysis}

Here we analyze the learned \texttt{LPP} policies in an effort to understand their representational complexity and qualitative behavior. We enumerate 10,000 programs and train with 11 task instances for each task. Table \ref{table:mapprograms} reports three statistics for the MAP policies learned for each task: the number of top-level programs (\texttt{at\_action\_cell} and \texttt{at\_cell\_with\_value} calls), the total number of method calls (\texttt{shifted}, \texttt{scanning}, and \texttt{cell\_is\_value} as well); and the maximum parse tree depth among all programs in the policy. The latter term is the primary driver of learning complexity in practice. The number of top-level programs ranges from 11 to 97; the number of method calls ranges from 32 to 244; and the maximum parse tree depth is 3 or 4 in all tasks. These numbers suggest that \texttt{LPP} learning is capable of discovering policies of considerable complexity, far more than what would be feasible with full policy enumeration. We also note that the learned policies are very sparse, given that the maximum number of available top-level programs is 10,000.   Visualizing the policies offers further confirmation of their strong performance (Figures 5--9). 

An example of a clause in the learned MAP policy for Stop the Fall is shown in Figure \ref{fig:stf_learned_clause}. This clause is used to select the first action: the cell below the parachuter and immediately above the floor is clicked. Other clauses govern the policy following this first action, either continuing to build blocks above the first one, or clicking the green button to turn on gravity. From the example clause, we see that the learned policy may include some redundant programs, which unnecessarily decrease the prior without increasing the likelihood. Such redundancy is likely due to the approximate nature of our Boolean learning algorithm (greedy decision-tree learning). Posthoc pruning of the policies or alternative Boolean learning methods could address this issue. The example clause also reveals that the learned policy may take unnecessary (albeit harmless) actions in the case where there are no ``fire'' objects to avoid. In examining the behavior of the learned policies for Stop the Fall and the other games, we do not find any unnecessary actions taken, but this does not preclude the possibility in general. Despite these two qualifications, the clause and others like it are a positive testament to the interpretability and intuitiveness of the learned policies.

\begin{figure*}[t]
    \centering
    \includegraphics[width=1.0\textwidth]{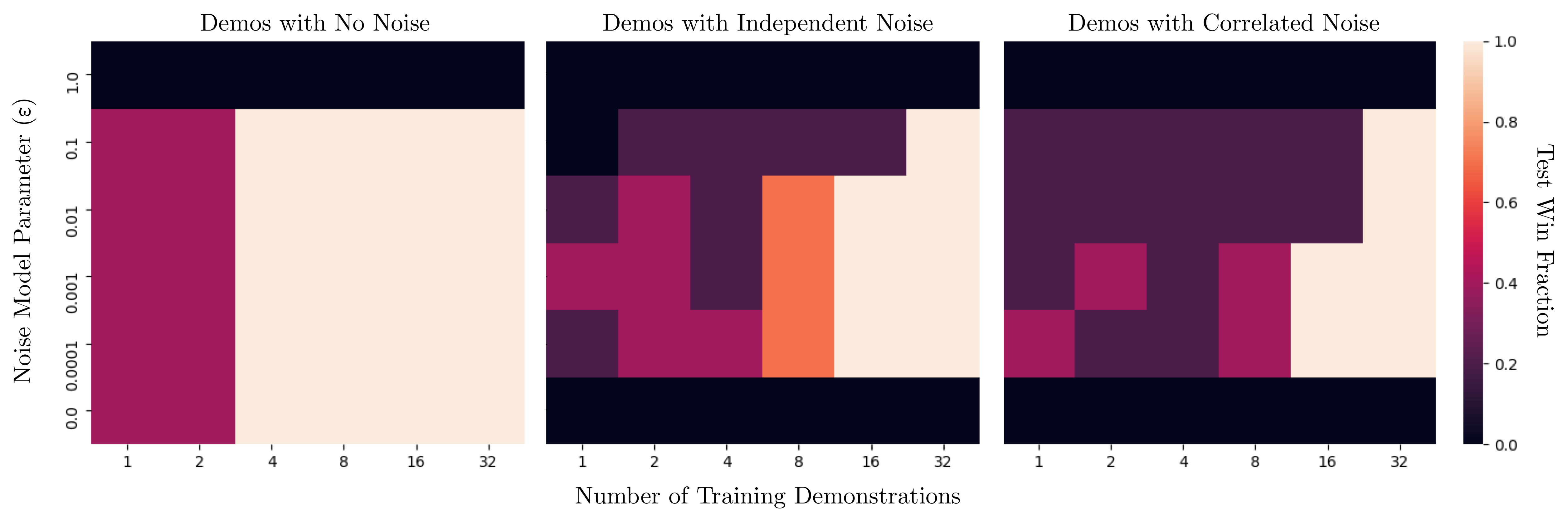}
    \caption{Effect of noisy demonstrations on learning performance and generalization in Nim. See text for details.}
    \label{fig:noisy_demonstrations}
\end{figure*}

\subsection{Noisy Demonstrations}

Our main experiments used demonstrations that were sampled following the expert policy exactly. 
In realistic settings, it is often the case that demonstrations are corrupted by noise, e.g., due to human error. 
We can accommodate noise in learning by modifying the likelihood $\mathsf{p}(\mathcal{D} \mid \pi)$ to include a noise model. 
A simple noise model that we consider here assumes that the expert policy is followed with probability $(1 - \epsilon)$, and with probability $\epsilon$, a random action is taken.
(This is akin to the expert following an $\epsilon$-greedy policy.)
The likelihood then becomes $\mathsf{p}(\mathcal{D} \mid \pi) \propto \prod_{i=1}^{N}\prod_{j=1}^{T_i} (1-\epsilon)\pi(a_{ij} \mid s_{ij}) + \frac{\epsilon}{|\mathcal{A}|}$.

In this set of experiments, we investigate how the noise model affects learning and generalization if demonstrations are (1) perfect, (2) corrupted by independent random noise, and (3) corrupted by correlated random noise. There will be mismatch between the noise model and noise source except for the case where the independent random noise probability of the demonstrations is equal to $\epsilon$ in the noise model.

\subsubsection{Demonstrations} We use the Nim task for all experiments. The perfect demonstrations (1) are identical to the main experiments. Independent random noise (2) is introduced by taking a random action with probability $0.2$. Correlated random noise (3) is introduced by clicking on a random \textit{token} cell with probability $0.2$. Recall that if a single incorrect token is clicked in Nim, the game is inevitably lost.

\subsubsection{Methods and Results} We modify the likelihood as described above and consider $\epsilon \in \{ 0.0, 1\mathrm{e}{-4}, 1\mathrm{e}{-3}, 1\mathrm{e}{-2}, 1\mathrm{e}{-1}, 1.0 \}$. All hyperparameters are otherwise unchanged from the main experiments. For each value of $\epsilon$, for each of the three sets of demonstrations, we perform \texttt{LPP} learning with demonstrations ranging from $2$ to $16$. 

Results are shown in Figure \ref{fig:noisy_demonstrations}. Across all demonstration types, we first note that too high of an $\epsilon$ results in a severe drop in performance. By inspecting the learned policies, we see that a high $\epsilon$ results in overly simplified policies that are favored by the prior; all actions not consistent with this simple policy are treated as noise. An $\epsilon$ of 0.0 also results in a severe drop in performance for the demos with noise, but for a different reason: when no noise is assumed by the model, no policy consistent with the noisy demos is found within the allotted enumeration budget. Interestingly, a very small but nonzero $\epsilon$ seems to be consistently effective, even though the actual probability of a random action in the noisy demos is $0.2$. This result suggests that the primary benefit of modifying the likelihood is to avoid discarding policy candidates that nearly, but not perfectly, match the demos.

% POINTS TO MAKE
% -0.0 and 1.0 don't work at all, but in between does. they fail for different reasons: 0.0 doesn't find any policies at all, and 1.0 finds too simple of a policy.
% -(MAYBE CHANGE CORRELATED NOISE SO THAT IT'S MORE DETRIMENTAL?)
% -we need more noisy demos than we did perfect demos, which makes sense. adding noise to the noise model doesn't affect the number of demos required to learn from perfect.
% -interestingly, it seems best to just add a very small epsilon. this has the effect of making imperfect decision trees have nonzero likelihood, so the algorithm doesn't fail completely.
% -and again if epsilon is too high then we'll learn too simple of a policy.
% -more sophisticated noise models are possible.

\subsection{Atari Breakout}

\begin{figure}[h]
    \centering
    \includegraphics[width=0.36\textwidth]{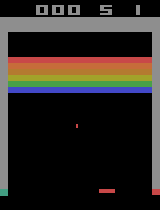}
    \caption{Atari Breakout.}
    \label{fig:breakout-frame}
\end{figure}

As our principal focus in this work is few-shot generalization, the experiments described thus far have been carried out in a suite of grid game tasks that were designed to exhibit substantial structured variation. Popular reinforcement learning benchmarks such as Atari 2600 games involve relatively little variation between task instances and were therefore not selected for our main experiments. Nonetheless, the question of whether $\texttt{LPP}$ policy learning and the particular DSL used for our grid games can be extended to Atari games and other familiar benchmarks is interesting, as it sheds light on the generality of the approach. In this section, we begin to answer this question with some preliminary experiments and results on Atari Breakout.

\subsubsection{Task}
In Breakout (Figure \ref{fig:breakout-frame}), an agent must move a paddle left or right so that it hits a ball into bricks above. When a brick is hit, it disappears, and a reward of 1 is given. If the ball falls below the paddle, the episode is over. The ball's initial position and velocity vary between task instances. We use the Breakout environment made available through OpenAI Gym (``BreakoutNoFrameskip-v4'') with three standard wrappers\footnote{These wrappers are implemented in the ``stable baselines'' package (\url{https://github.com/hill-a/stable-baselines}). Equivalent preprocessing was done in the original DQN paper and downstream work.}: ``EpisodicLifeEnv'', which ensures that the episode ends when the ball falls below the paddle; ``FireResetEnv'', which executes the ``fire'' action at the beginning of the game to make the ball appear; and ``NoopResetEnv'', which randomizes the initial position of the ball between task instances. Raw observations are $210 \times 160$ RGB images. We stack two consecutive observations to capture velocity information, resulting in size $210 \times 160 \times 3 \times 2$ tensors\footnote{Stacking four frames is standard for Atari but redundant for Breakout}. We convert these tensors into the familiar 2D grid representation by flattening the latter two dimensions so that each $3 \times 2$ tensor is converted to a single float value. In the data, there are 15 unique flattened values across all pixels in all observations. The final observation space is thus isomorphic to a 2D grid with 15 discrete values, that is, $\{0, 1, ..., 14\}^{210 \times 160}$.

%As only the ball, paddle, and bricks change, there are a total of 15 distinct cell values. The final observation is thus in $\{0, 1, ..., 14\}^{210 \times 160}$.

\subsubsection{Expert Policy}
We implement a simple reactive policy from which we can draw expert demonstrations. Given an observation, we first determine the velocity of the ball based on the two consecutive frames. We further determine whether the paddle is to the left or right of the ball. If the ball is moving towards the left and the paddle is not on the left, then we take the ``left'' action, and the same for the opposite direction. We take the noop action otherwise. This policy is not necessarily optimal, but it successfully keeps the ball in play and continues to reap rewards for several thousand frames.

\subsubsection{Methods and Results}

We use the same DSL described in the main paper for this experiment, with one change: the \texttt{at\_action\_cell} method of the DSL is designed under the assumption that actions are clicks on a grid cell, but the action space in Atari Breakout is simply four discrete values (``left'', ``right'', ``fire'', ``noop''); thus we remove this method from the DSL and replace it with four simple identity checks (\texttt{action\_is\_left, action\_is\_right, action\_is\_fire, action\_is\_noop}). The probabilistic grammatical prior is modified accordingly, giving uniform probability to these four action methods and to the \texttt{at\_cell\_with\_value} method in the first substitution.

We sample a single demonstration for 500 frames using the expert policy described above. This demonstration shows 3 bricks breaking for a total reward of 3. We then run $\texttt{LPP}$ learning with 120,000 programs enumerated. (Fewer programs were found to be insufficient.) To ease the computational burden, we enumerate programs involving only 4 of the 15 total discrete values, namely those involving the color of the paddle and ball. This is a problem-specific hack that injects additional prior knowledge. However, we do expect that results would be similar if all discrete values were included and more programs (on the order of 5 million) were enumerated.

To evaluate the learned policy, we test on 10 held-out task instances for 500 frames. (Recall that variation between task instances is parameterized by the initial position and velocity of the ball.) In all 10 cases, the learned policy accrues a reward of 3. This matches the expert policy and suggests perfect generalization from only a single relatively short demonstration, at least for the limited task horizon of 500. To further assess generalization, we continue past the horizon of 500 and record the total reward accumulated before the ball falls below the paddle. Across the 10 held-out task instances, we find an average reward of 8.80 with a standard deviation of 7.47. These results indicate that the learned policy can generalize immediately to observations with more bricks missing in different locations than seen during the single demonstration.

These preliminary results suggest that the policy learning method and the specific DSL proposed in this work may extend to other tasks with minor modification, including games that are not typically thought of as grid-based. We look forward to continuing to evaluating the scalability and applicability of the DSL and general \texttt{LPP} policy learning in future work.

\subsection{Training and Test Environments}

In the following figures, we illustrate each task with representative training demonstrations and \texttt{LPP} test performance. (See the main text for Fence In.)

\begin{figure*}[h]
    \centering
    \includegraphics[width=1.0\textwidth]{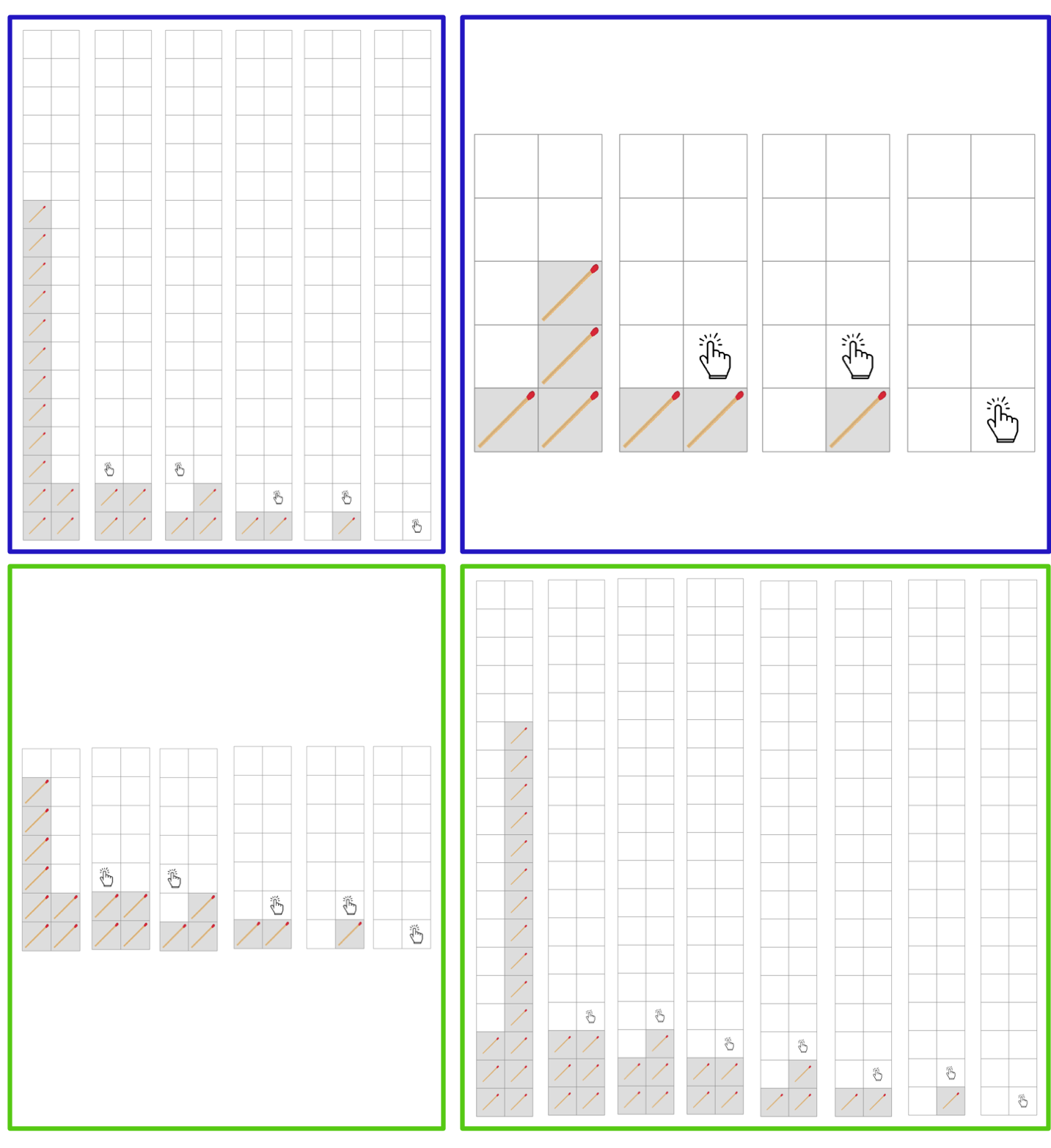}
    \caption{A \texttt{LPP} policy learned from two demonstrations of ``Nim'' (blue) generalizes perfectly to all test task instances (e.g. green).}
    \label{fig:nim_panels}
\end{figure*}
\clearpage

\begin{figure*}[h]
    \centering
    \includegraphics[width=1.0\textwidth]{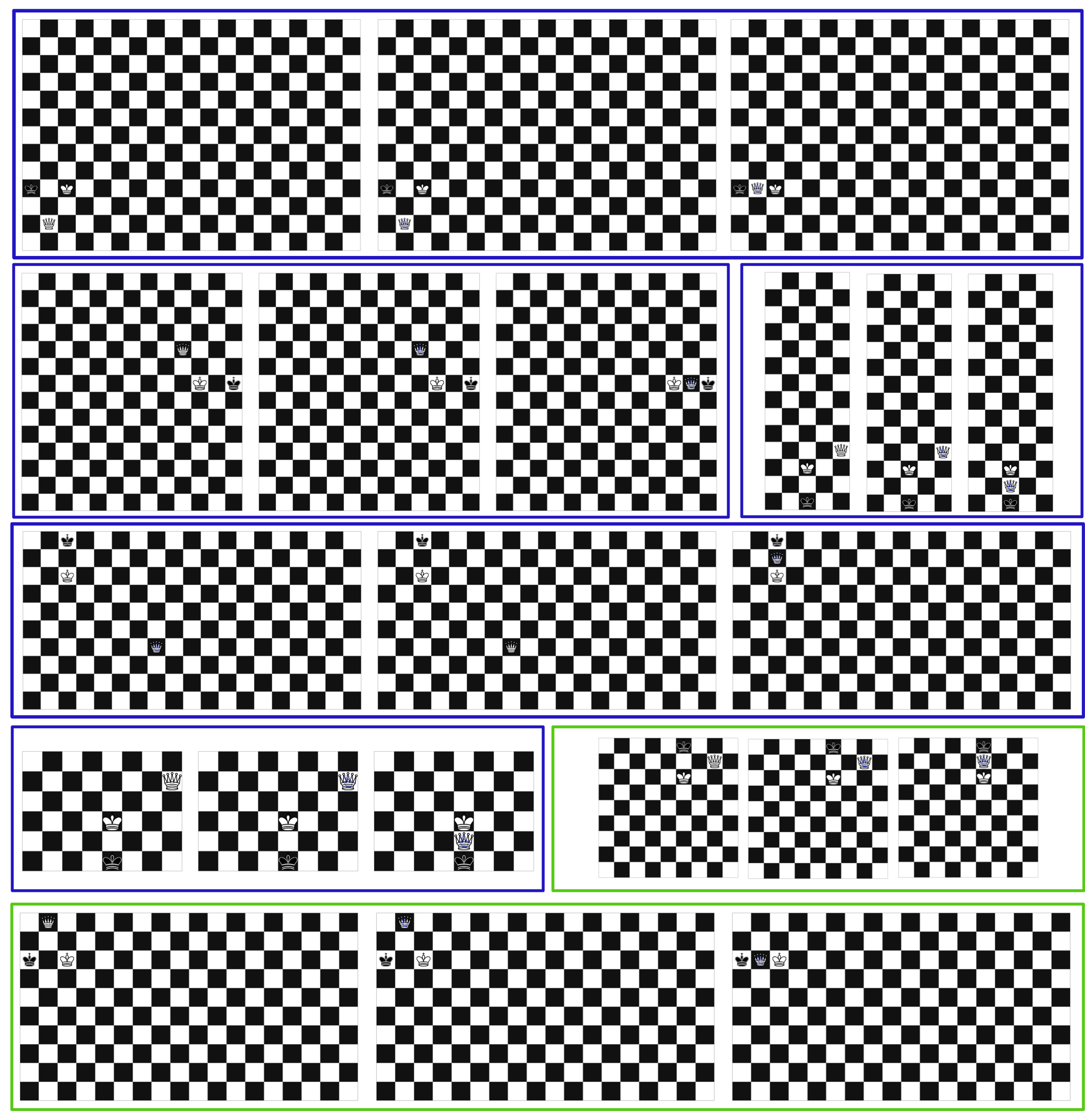}
    \caption{A \texttt{LPP} policy learned from five demonstrations of ``Checkmate Tactic'' (blue) generalizes perfectly to all test task instances (e.g. green).}
    \label{fig:chess_panels}
\end{figure*}
\clearpage

\begin{figure*}[h]
    \centering
    \includegraphics[width=0.8\textwidth]{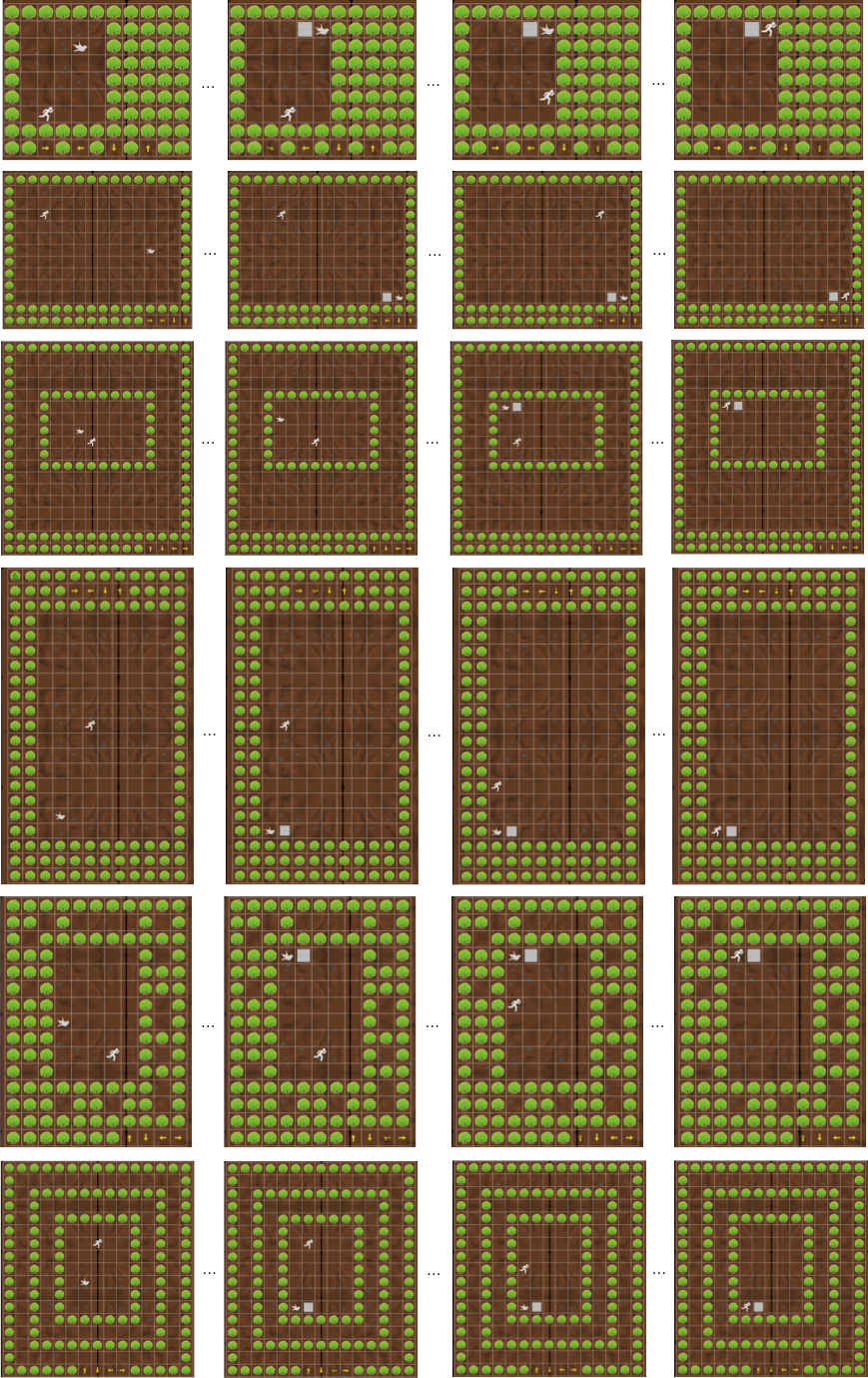}
    \caption{A \texttt{LPP} policy learned from four demonstrations of ``Chase'' (top) generalizes perfectly to all test task instances (e.g. bottom).}
    \label{fig:chase_panels}
\end{figure*}
\clearpage

\begin{figure*}[h]
    \centering
    \includegraphics[width=1.0\textwidth]{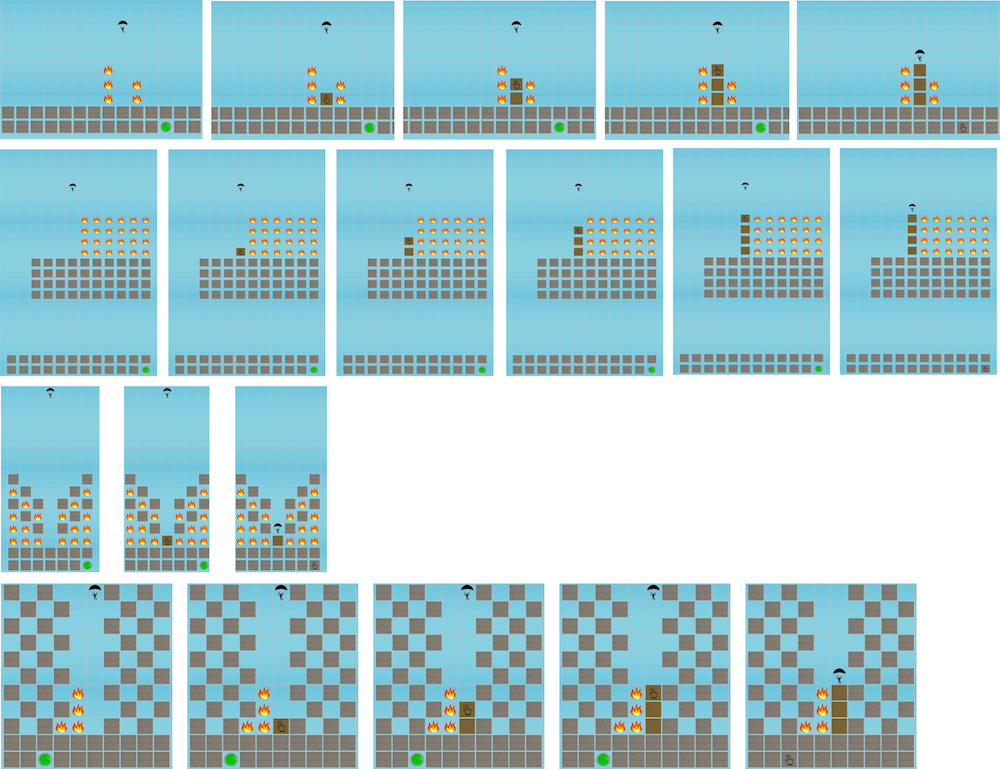}
    \caption{A \texttt{LPP} policy learned from three demonstrations of ``Stop the Fall'' (top) generalizes perfectly to all test task instances (e.g. bottom).}
    \label{fig:stf_panels}
\end{figure*}
\clearpage

\begin{figure*}[h]
    \centering
    \includegraphics[width=0.9\textwidth]{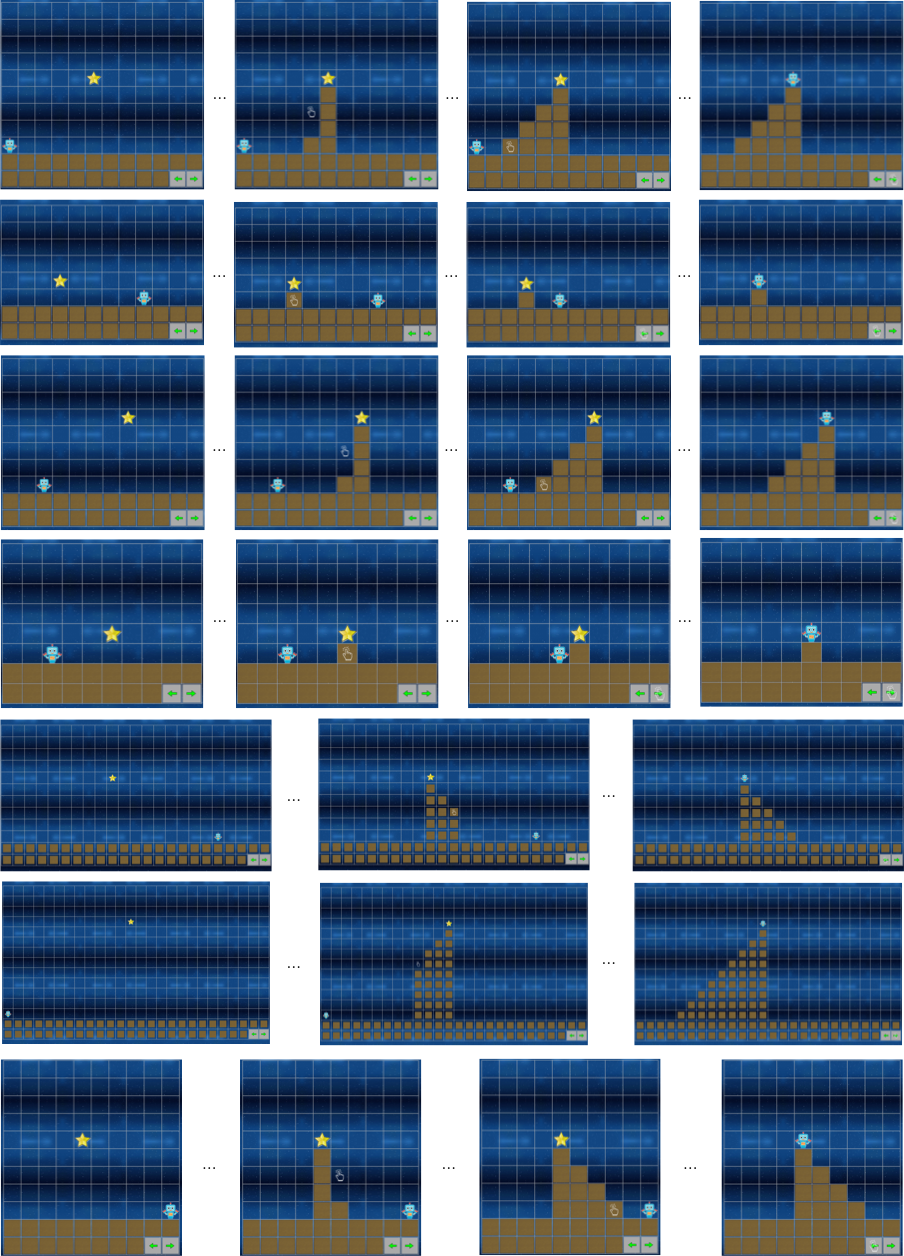}
    \caption{A \texttt{LPP} policy learned from five demonstrations of ``Reach for the Star'' (top) generalizes perfectly to all test task instances (e.g. bottom).}
    \label{fig:rfts_panels}
\end{figure*}
\clearpage

\end{document}